%% file: main.tex
\documentclass[12pt]{article} 
\usepackage{aaai25}  
\usepackage{times}  
\usepackage{helvet}  
\usepackage{courier}  
\usepackage[hyphens]{url}  
\usepackage{graphicx} 
\usepackage{natbib}  
\usepackage{caption} 
\usepackage{algorithm}
\usepackage{algorithmic}

\usepackage{newfloat}
\usepackage{float}
\usepackage{listings}
\DeclareCaptionStyle{ruled}{labelfont=normalfont,labelsep=colon,strut=off} 
\floatstyle{ruled}
\newfloat{listing}{tb}{lst}{}
\floatname{listing}{Listing}
\usepackage{xspace}
\newcommand{\github}{\raisebox{-1.5pt}{\includegraphics[height=1.3em]{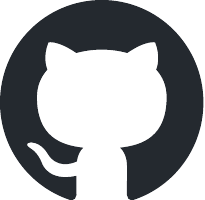}}\xspace}
\usepackage{libertinus}
\usepackage{graphicx}
\usepackage{makecell}
\usepackage{multirow}
\usepackage{multicol}
\usepackage{makecell}
\usepackage{booktabs}
\usepackage{colortbl}
\usepackage{array}
\usepackage{url}
\usepackage{fontawesome}
\usepackage{enumitem}
\usepackage{amsmath}
\usepackage{amssymb}
\usepackage[table,xcdraw]{xcolor}
\usepackage{multirow}
\usepackage{makecell}
\usepackage{pifont}
\usepackage{arydshln}
\usepackage{todonotes}
\usepackage{fontawesome}
\usepackage{bm}
\usepackage{booktabs}
\usepackage[colorlinks=true, linkcolor=blue, citecolor=blue, urlcolor=blue]{hyperref}

\usepackage{wrapfig}

\title{\raisebox{-0.8ex}{\includegraphics[height=0.35in, width=0.35in]{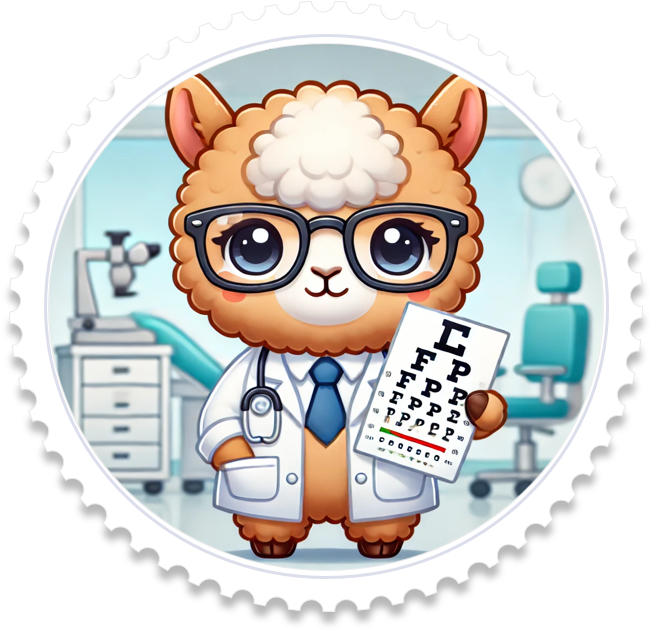}}EyecareGPT: Boosting Comprehensive Ophthalmology
Understanding with Tailored Dataset, Benchmark and Model}

\author {
    \small
    Sijing Li\textsuperscript{\rm 1},
    Tianwei Lin\textsuperscript{\rm 1},
    Lingshuai Lin\textsuperscript{\rm 2},
    Wenqiao Zhang\textsuperscript{\rm 1},
    Jiang Liu\textsuperscript{\rm 1},
    Xiaoda Yang\textsuperscript{\rm 1},
    Juncheng Li\textsuperscript{\rm 1},\\
    Yucheng He\textsuperscript{\rm 3},
    Xiaohui Song\textsuperscript{\rm 1},
    Jun Xiao\textsuperscript{\rm 1},
    Yueting Zhuang\textsuperscript{\rm 1},
    Beng Chin Ooi\textsuperscript{\rm 4}
}
\affiliations {
    \small
    \textsuperscript{\rm 1}Zhejiang University,
    \textsuperscript{\rm 2}Harbin Institute of Technology,
    \textsuperscript{\rm 3}The First People's Hospital of Chenzhou,\\
    \textsuperscript{\rm 4}National University of Singapore\\
    \vspace{2mm}
    {\github \href{https://github.com/DCDmllm/EyecareGPT}{{\text{\, Code}}}}
}

\begin{document}
\twocolumn[
    {
        \renewcommand\twocolumn[1][]{#1}
        \vspace{-12mm}
        \maketitle
        \vspace{-3mm}
        \begin{center}
        \captionsetup{type=figure}
        \includegraphics[width=0.98\textwidth]{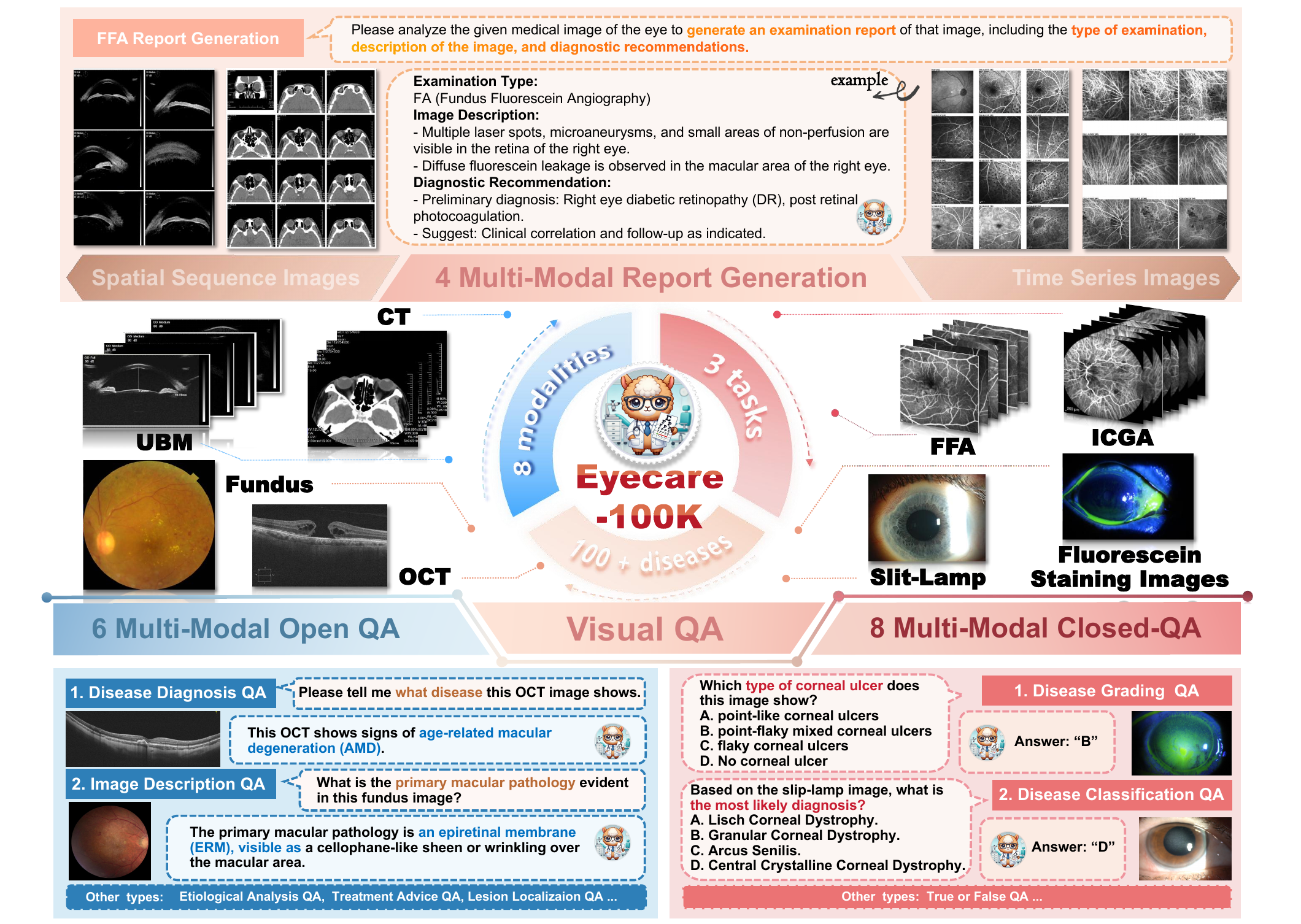}
        \captionof{figure}{\textbf{Overview of the Eyecare-100K.} Eyecare-100K aggregates real-world ophthalmic data across \textbf{8 modalities, 15+ anatomical structures and 100+ eye diseases}, supporting multi-modal report generation and fine-grained visual QA tasks. }
        \label{fig:1}
        \end{center}
    }
]
\input{tex/0abstract}
\input{tex/1introduction}

\input{tex/2relatedwork}
\input{tex/3dataset}
\input{tex/4benchmark}
\input{tex/5method}

\input{tex/6experiments}

\input{tex/7conclusion}
\nocite{zhang2024hyperllava}
\bibliography{ref.bib}

\newpage
\appendix
\onecolumn
\input{tex/Appendix}

\end{document}

%% file: tex/0abstract.tex
\begin{abstract}

Medical Large Vision-Language Models (Med-LVLMs) demonstrate significant potential in healthcare, but their reliance on general medical data and coarse-grained global visual understanding limits them in intelligent ophthalmic diagnosis. 
Currently, intelligent ophthalmic diagnosis faces three major challenges: \textbf{(i) Data.} The lack of deeply annotated, high-quality, multi-modal ophthalmic visual instruction data;
\textbf{(ii) Benchmark.} The absence of a comprehensive and systematic benchmark for evaluating diagnostic performance;
\textbf{(iii) Model.} The difficulty of adapting holistic visual architectures to fine-grained, region-specific ophthalmic lesion identification. 
In this paper, we propose the \textbf{Eyecare Kit}, which systematically tackles the aforementioned three key challenges with the tailored dataset, benchmark and model: First, we construct a multi-agent data engine with real-life ophthalmology data to produce \textbf{Eyecare-100K}, a high-quality ophthalmic visual instruction dataset. 
Subsequently, we design \textbf{Eyecare-Bench}, a benchmark that comprehensively evaluates the overall performance of LVLMs on intelligent ophthalmic diagnosis tasks across multiple dimensions. 
Finally, we develop the \textbf{EyecareGPT}, optimized for fine-grained ophthalmic visual understanding thoroughly, which incorporates an adaptive resolution mechanism and a layer-wise dense connector. 
Extensive experimental results indicate that the EyecareGPT achieves state-of-the-art performance in a range of ophthalmic tasks, underscoring its significant potential for the advancement of open research in intelligent ophthalmic diagnosis. Our project is available at \url{https://github.com/DCDmllm/EyecareGPT}.\looseness=-1
\end{abstract}

%% file: tex/1introduction.tex
\section{Introduction}
Large Vision-Language Models (LVLMs)~\cite{liu2024llavanext,liu2024improved,lin2024vila,team2024gemini,hurst2024gpt} achieve remarkable progress in open-world visual understanding tasks~\cite{ren2024grounded,zhang2019frame}, demonstrating potential in medical scenarios. In recent years, Medical Large Vision-Language Models (Med-LVLMs), such as LLaVA-Med~\cite{li2023llava}, HuatuoGPT-Vision\cite{chen2024huatuogpt}, and HealthGPT~\cite{lin2025healthgpt}
trained on extensive medical visual instruction data, advanced medical tasks including pathological diagnosis and knowledge reasoning. 
However, due to the lack of deep modeling of modality-specific features and domain-specific expertise in vertical medical fields, existing Med-LVLMs still exhibit significant limitations in fine-grained visual understanding and region-specific intelligent diagnosis within specific disciplines. 
Taking ophthalmology as an example, the field involves complex medical imaging modalities and highly specialized clinical requirements~\cite{balas2024adaptive,xu2024unveiling}, where current ophthalmic foundation models~\cite{shi2024eyeclip} and Med-LVLMs~\cite{saab2024capabilities,alsaad2024multimodal,lin2025healthgpt} fail to provide effective support. Therefore, developing advanced Med-LVLM specifically tailored for ophthalmology, equipped with fine-grained visual understanding and reasoning capabilities, becomes an urgent need to drive intelligent ophthalmology research and applications.\looseness=-1

Effectively transferring the Med-LVLM paradigm to ophthalmology requires a systematic analysis of the domain's unique requirements and challenges.
First, existing ophthalmic datasets primarily focus on single imaging modality and pathological classification task, lacking deeply annotated, high-quality, heterogeneous multi-modal visual instruction data that cover temporal sequence data (e.g., FFA, ICGA), spatial sequence data (e.g., UBM, CT), and complex modalities (e.g., Fundus, OCT, Slit-Lamp). 
Second, the absence of a comprehensive benchmark for intelligent ophthalmic diagnosis hinders accurate evaluation of Med-LVLMs on fine-grained visual understanding and report generation tasks, and limits guidance for model optimization.
Finally, current Med-LVLM architectures, relying on coarse-grained global features, often overlook critical fine-grained priors and local details in ophthalmic imaging, failing to meet the precision requirements of intelligent diagnosis.
To address these challenges, we propose the \textbf{Eyecare Kit}, which systematically advances the adaptability and performance of Med-LVLMs in ophthalmology through innovations in three key aspects: \textbf{Dataset}, \textbf{Benchmark}, and \textbf{Model}.\looseness=-1

\textbf{(i) Dataset.} 
To address the scale, modality, and task diversity of ophthalmic data, we collect real-world data from 13 public datasets, 3 hospitals, and 3 public medical case libraries, covering 8 imaging modalities, over 15 anatomical structures, and more than 100 types of eye diseases (Figure~\ref{fig:1}). 
A \textbf{multi-agent data engine} is developed for information extraction, cleaning, formatting, and expert review, resulting in a comprehensive dataset named \textbf{Eyecare-100K}, containing approximately 102,000 visual question answering (VQA) pairs. 
Eyecare-100K is the first comprehensive ophthalmic dataset to simultaneously encompass multiple modalities, tasks, and diseases, and is expected to serve as a key resource for advancing multi-modal intelligent understanding in ophthalmology.\looseness=-1

\textbf{(ii) Benchmark.} 
To deeply evaluate the comprehensive ophthalmology understanding capability of a Med-LVLM, we develop a benchmark named \textbf{Eyecare-Bench}, which includes three key clinical metrics: closed QA, open QA, and report generation. The data instances in Eyecare-Bench are drawn from the designated test set of Eyecare-100K, comprising about 15,000 carefully sampled examples across tasks, modalities, and disease categories to ensure balanced and representative evaluation. Notably, we design multi-dimensional evaluation metrics for different tasks and introduce GPT-4 to provide a more comprehensive evaluation of the report generation capabilities of models. Eyecare-Bench provides significant reference value for the open community to research more accurate and reliable intelligent systems for eye diseases.

\textbf{(iii) Model.} 
To address the fine-grained and region-specific demands of intelligent ophthalmic diagnosis and to validate the effectiveness of the Eyecare Kit in supporting high-quality data and systematic evaluation, we propose the \textbf{EyecareGPT} model. We employ SigLIP~\cite{zhai2023sigmoid} as a high-resolution visual feature extractor to enhance local lesion perception. To accommodate variable resolutions in clinical ophthalmic imaging, we design an adaptive resolution mechanism~\cite{you2024ferret,guo2025llava,zhang2022boostmis} for dynamic adjustment, improving consistency across multi-modal images. Additionally, we introduce a Layer-wise Dense Connector (LDC) to densely integrate multi-scale visual features and preserve fine-grained structural information. Finally, we provide two scalable EyecareGPT variants to enable flexible deployment in diverse real-world settings.

Experimental results show that the Eyecare Kit provides high-quality, deeply annotated data for intelligent ophthalmic diagnosis and establishes a comprehensive evaluation benchmark, effectively supporting the optimization and development of Med-LVLMs. The main contributions of this work are as follows:

\begin{itemize}[leftmargin=1em, itemsep=0em, parsep=0em, topsep=0em]
\item \textbf{High-Quality Dataset.} We propose the first comprehensive ophthalmic visual instruction dataset.
\item \textbf{Comprehensive Benchmark.} We build a systematic benchmark to evaluate the clinical performance of LVLMs on 3 core tasks: closed QA, open QA and report generation.
\item \textbf{Adapted LVLM Architecture.} We introduce an LVLM architecture adapted to complex, heterogeneous ophthalmic clinical imaging, achieving SoTA performance.
\item \textbf{Facilitating Open Research.} We will fully open-source the dataset, benchmark, and model to facilitate research on intelligent ophthalmic diagnosis.
\end{itemize}

\vspace{-2mm}

%% file: tex/2relatedwork.tex
\section{Related Work}
\subsection{Medical Large-Vision Language Models}
Med-LVLMs achieve groundbreaking progress in processing and understanding medical imaging, offering new possibilities for clinical diagnosis and treatment~\cite{chen2024towards,xu2024mlevlm}. Med-Flamingo~\cite{moor2023med} leverages multimodal knowledge across medical disciplines for pre-training, extending the Flamingo~\cite{alayrac2022flamingovisuallanguagemodel} framework into the medical domain. Models such as LLaVA-Med~\cite{li2023llava} and UMIT~\cite{yu2025umit} adopt a two-stage training strategy combining pre-training and fine-tuning, enhancing vision-text alignment and multitask adaptability. To address language adaptability and dataset specificity challenges, HuatuoGPT-Vision~\cite{chen2024huatuogpt} introduces the PubMedVision dataset, comprising 1.3 million high-quality medical samples and markedly improving model adaptability. Specialized LVLMs like LLava-Rad~\cite{zambrano2025clinically} focus on radiology image understanding, actively exploring report generation tasks aligned with clinical practice. However, existing specialized ophthalmic models exhibit limited generalization; for instance, DeepDR-LLM~\cite{li2024integrated} is restricted to auxiliary diagnosis of diabetic retinopathy, and Ophtha-LLaMA2~\cite{zhao2023ophthallama2largelanguagemodel} uses only three ophthalmic modalities for fine-tuning. Currently, Med-LVLMs are gradually evolving from general medical tasks toward clinical practicality, yet the scarcity of high-quality datasets in specific subfields continues to hinder their development.
\subsection{Ophthalmic Multi-Modal Datasets}
High-quality ophthalmic datasets hold significant clinical and societal value in advancing intelligent diagnostic models. Currently, publicly available datasets primarily consist of ocular images labeled with classification tags or diagnostic keywords. For ocular disease classification, typical fundus image datasets include IDRID~\cite{porwal2018indian}, ACRIMA~\cite{ovreiu2021deep}, JSIEC~\cite{cen2021automatic}, ODIR~\cite{ODIR2019}, Harvard-GDP~\cite{luo2023harvard}, MuRed~\cite{rodriguez2022multi}, and DeepDRiD~\cite{liu2022deepdrid}. Similarly, OCT2017~\cite{kermany2018labeled}, Kermany~\cite{kermany2018oct}, OCTID~\cite{gholami2020octid}, and OCTDL~\cite{kulyabin2024octdl} provide ocular OCT images with corresponding classification labels. Although these datasets contribute significantly to their respective tasks, they exhibit notable limitations in imaging modality diversity and fine-grained annotation, restricting their use in more complex intelligent diagnostic applications. The recently proposed multimodal ophthalmic dataset LMOD~\cite{qin2025lmodlargemultimodalophthalmology} covers five imaging modalities, partially addressing the limitations of single-modality datasets. However, the lack of medical visual instruction datasets for LVLM training highlights the need to develop larger, more diverse, and finely annotated heterogeneous multimodal ophthalmic datasets to support model training and evaluation.

%% file: tex/3dataset.tex
\section{Eyecare Kit: Eyecare-100K}
\subsection{Data Collation and Organization}
Existing ophthalmic datasets are typically limited to a single task or modality and suffer from inconsistent standards and uneven distributions. These limitations make them unsuitable for constructing high-quality visual instruction datasets, thereby restricting the development of Med-LVLMs in intelligent ophthalmic applications. To address this, we propose \textbf{Eyecare-100K}, a comprehensive ophthalmic visual instruction dataset that covers diverse heterogeneous multi-modal imaging, aiming to provide standardized data support for intelligent understanding in ophthalmology.\looseness=-1

\begin{figure}[t]
    \centering
    \includegraphics[width=\linewidth]{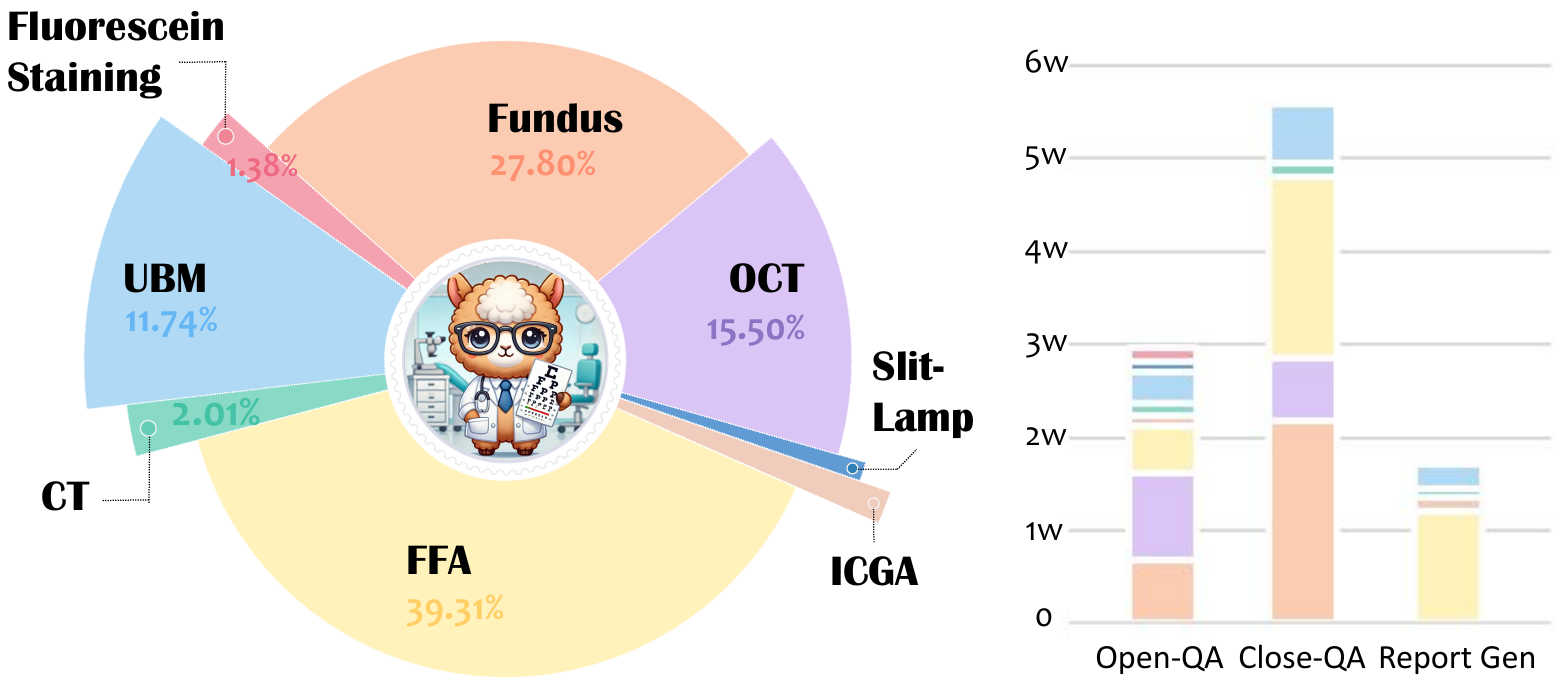}
    \caption{Data statistics of {\sffamily\textbf{Eyecare-100K}}. }
    \label{fig:data}
    \vskip -0.12in
\end{figure}

To address the lack of available data for key clinical modalities, we collaborate with three large public hospitals to collect and annotate real-world ophthalmic cases with anonymization procedures. We also systematically collect and standardize multiple public ophthalmic datasets across different modalities and labels (see Appendix for details). To further expand diversity, Eyecare-100K incorporates examination cases from public medical repositories such as Radiopaedia~\cite{radiopaedia} and MedPix~\cite{siragusa2024medpix}, as well as professional slit-lamp textbooks~\cite{liang2022qianyanjie}. In total, Eyecare-100K integrates 58,485 ophthalmic images from 13 public datasets, 3 hospitals, and 3 medical case repositories. The dataset covers 8 imaging modalities (see Fig.~\ref{fig:data}): \textbf{(1)} Fluorescein Angiography (FA), \textbf{(2)} Indocyanine Green Angiography (ICGA), \textbf{(3)} Optical Coherence Tomography (OCT), \textbf{(4)} Fundus Photography, \textbf{(5)} Ultrasound Biomicroscopy (UBM), \textbf{(6)} Slit-Lamp, \textbf{(7)} Fluorescein Staining Imaging, and \textbf{(8)} Computed Tomography (CT), spanning 15 anatomical structures and over 100 ophthalmic diseases and rare conditions, significantly enhancing dataset diversity and comprehensiveness.\looseness=-1

Considering that traditional medical data primarily consist of classification labels, segmentation annotations, or brief textual descriptions and lack the visual instruction structures needed for fine-tuning Med-LVLMs, we develop a multi-agent data engine to extract, clean, standardize, and perform expert review on large-scale raw data. Ultimately, the data are organized into three types of VQA tasks: closed QA (multiple-choice questions), open QA (short-form questions), and report generation (long-text answers), to equip the models with fine-grained ophthalmic visual understanding and reasoning capabilities.\looseness=-1
\subsection{Multi-Agent Data Engine}
\begin{figure}[t]
    \centering
    \includegraphics[width=0.99\linewidth]{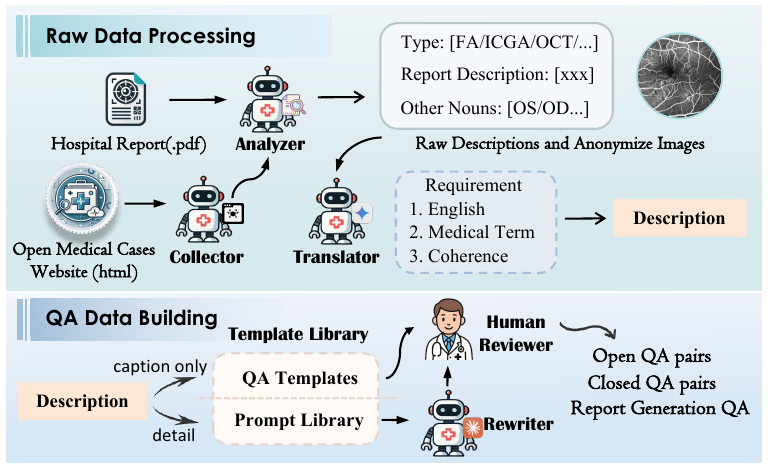}
    \caption{Framework of Multi-Agent Data Engine.}
    \label{fig:agent}
    \vskip -0.2in
\end{figure}

We develop an automated \textbf{multi-agent data engine} to create Eyecare-100K, converting categorized labels and raw reports into structured VQA pairs. As shown in Figure~\ref{fig:agent}, the agent engine comprises 6 components as follows.

\noindent \textbf{{Analyzer for Description Extraction.}} 
Given that a large volume of raw clinical data is stored in PDF format, containing imaging modalities, diagnostic results, and other details. Qwen2-VL-2B-OCR~\cite{Qwen2VL} is adapted as our analyzer to automatically extract key information and anonymize sensitive information. 

\noindent \textbf{{Intelligent Collector for Medical Cases.}} 
We construct an intelligent collector to extract metadata from authorized public medical case repositories. The extracted metadata includes imaging modalities, anatomical structures, and diagnostic descriptions, enriching the diversity of data sources and expanding the coverage of clinical cases.

\noindent \textbf{{Translator for Data Sanitization.}} 
To address the common issues of complex abbreviations and mixed-language expressions in medical descriptions, we integrate Gemini-2.0-Flash~\cite{gemini2.0flash} as an automated translator. This module accurately interprets the contextual meaning of medical terms and converts raw descriptions into clear, standardized professional language, thereby enhancing the consistency of the dataset.

\noindent \textbf{{Template Library of QA and Prompt.}} 
To support fine-grained tasks (closed QA, open QA, report generation), we designed a diverse VQA and prompt template library (see Appendix). For single-label data, we apply VQA templates to create open tasks involving modality identification, disease diagnosis, and lesion grading. For data containing detailed pathological information, we use prompt templates to generate all three types of tasks. This design facilitates model training across multiple dimensions—understanding, reasoning, and generation—enhancing overall performance in real-world clinical scenarios.

\noindent \textbf{{Rewriter for Generating VQAs.}} 
Claude-3.7 serves as the rewriter, using prompts to extract key information from processed descriptions and construct reliable VQA pairs with the extracted information as answers. It supports generating VQA types such as modality identification, disease diagnosis and lesion localization, etc. For report generation prompts, it automatically organizes three components—examination types, imaging findings, and diagnostic recommendations—from the processed descriptions and generates a standardized Markdown format report.

\noindent \textbf{{Human Preference-Based Reviewer.}}
To improve the accuracy and standardization of automatically generated data, we randomly select 10\% constructed VQA instructions and report samples, and introduce five human reviewers to inspect the data quality. Medical professionals concentrate on semantic validity, medical consistency, and format standardization.
Each data entry undergoes two rounds of independent review, effectively ensuring the fine-grained data quality of Eyecare-100K.

%% file: tex/4benchmark.tex
\section{Eyecare Kit: Eyecare-Bench}
We propose \textbf{Eyecare-Bench} to systematically evaluate the performance of Med-LVLMs in intelligent ophthalmic diagnosis. The data instances in Eyecare-Bench are drawn from the designated test set of Eyecare-100K, comprising about 15,000 examples. Sampling is carefully designed across all task types, imaging modalities, and ophthalmic disease categories to ensure representative proportions within each class, maintaining the balance and comprehensiveness of the test set. To the best of our knowledge, Eyecare-Bench is the most comprehensive benchmark to date for evaluating LVLMs in ophthalmology.

\input{table/metric_table}
\subsection{Multi-Dimensional Evaluation Suite}
To systematically evaluate model performance on multi-task and multi-format ophthalmic VQA tasks, we design a multi-dimensional evaluation suite, \textbf{EyeEval}. EyeEval defines fine-grained evaluation metrics for three task types, comprehensively covering aspects such as generation quality, factual consistency, and linguistic faithfulness.

\noindent \textbf{{VQA Evaluation Metrics}}
In closed QA tasks, questions are presented in a multiple-choice format, aiming to assess the response accuracy of models. Therefore, we adopt accuracy as the primary evaluation metric. For open QA tasks, we focus on the factual consistency and linguistic coherence of generated answers. Specifically, we use F1-RadGraph~\cite{yu2023evaluating}, BERTScore-F1~\cite{zhang2019bertscore}, and F1-BioBert~\cite{lee2020biobert} to evaluate factual accuracy, BLEU~\cite{papineni2002bleu} and ROUGE-L~\cite{lin2003automatic} to assess surface similarity and language faithfulness.

\noindent \textbf{{Report Generation Evaluation Metrics.}} 
We recognize that traditional evaluation methods based on n-grams~\cite{culy2003limits} or semantic embeddings~\cite{bakarov2018survey} often introduce bias due to the inherent diversity and uncertainty of long-form text generation and correlate poorly with expert assessments. Therefore, in addition to the commonly used NLP metrics, we find five authoritative experts to develop a Ten-criteria evaluation framework (see Table~\ref{tab:metric_table}) covering four key dimensions: \textbf{accuracy}, \textbf{completeness}, \textbf{structural rationality}, and \textbf{clinical practicability}. Each indicator is assigned a weight according to its clinical importance, and the total score of report is capped at 100. Evaluations are conducted using GPT-4 based on this refined rubric.

Specifically, indicators \textbf{(A)–(D)} are quantitative metrics assessing the accuracy and completeness of abnormal findings in the report. The next five are Boolean indicators: \textbf{(E)–(H)} evaluate the structural coherence of the report, and \textbf{(I)} assesses the presence of critical errors that could affect clinical decision-making. Indicator \textbf{(J)} independently evaluates the diagnostic accuracy of the report. For quantitative indicators, deductions are applied proportionally based on the number of errors and corresponding weights. For Boolean indicators, points are deducted if the condition is not met.

According to the scoring criteria, we grade the reports as follows: 
\textbf{(i) Excellent Report (90-100).}
\textbf{(ii) Usable Report (80-90).}
\textbf{(iii) Report Under Review (60-80).} 
\textbf{(iv) Unusable Report (Below 60).}
In the subsequent experiments, we validate the consistency between this evaluation framework and human expert assessments, thereby demonstrating the reliability of the framework and clinical relevance in quantitatively measuring report quality.

%% file: table/metric_table.tex
\begin{table}[t]
\centering
\caption{Ten-Criteria evaluation framework.}
\vskip -0.12in
\resizebox{\columnwidth}{!}{%
\begin{tabular}{cc}
\midrule[1.5pt]
{\textbf{Definition of Indicators}} & {\textbf{Weight}} \\
\midrule[0.5pt]
\rowcolor{gray!5} \multicolumn{2}{c}{\textbf{Scoring Indicators}} \\
\midrule[0.5pt]
\makecell[l]{\parbox[t]{10cm}{A) The number of abnormal features in candidate report that are not mentioned in the reference report.}}       & 1  \\
\midrule[0.5pt]
\makecell[l]{\parbox[t]{10cm}{B) The number of times the candidate report describes the disease severity incorrectly.}}          & 4  \\
\midrule[0.5pt]
\makecell[l]{\parbox[t]{10cm}{C) The number of times the candidate report describes the disease location incorrectly.}}          & 4  \\
\midrule[0.5pt]
\makecell[l]{\parbox[t]{10cm}{D) The number of missing key findings compared to the reference report.}}                                      & 6  \\
\midrule[0.5pt]
\makecell[l]{\parbox[t]{10cm}{E) Whether the diagnosis or suspected diagnosis is included.}}                                                 & 2  \\
\midrule[0.5pt]
\makecell[l]{\parbox[t]{10cm}{F) Whether the description of the examination type exists and is correct.}}                                   & 2  \\
\midrule[0.5pt]
\makecell[l]{\parbox[t]{10cm}{G) Whether there is a treatment recommendation.}}                                                              & 2  \\
\midrule[0.5pt]
\makecell[l]{\parbox[t]{10cm}{H) Whether the report structure is clear.}}                                                                    & 5  \\
\midrule[0.5pt]
\makecell[l]{\parbox[t]{10cm}{I) Whether the candidate outcome contains particularly serious clinical errors.}}                              & 15 \\
\midrule[0.5pt]
\rowcolor{gray!5} \multicolumn{2}{c}{\textbf{Correct Rate Calculation Indicator}} \\
\midrule[0.5pt]
\makecell[l]{\parbox[t]{10cm}{J) Whether the diagnosis is similar or approximately correct.}}                                                & -  \\
\midrule[1.5pt]
\end{tabular}
}
\vskip -0.18in
\label{tab:metric_table}
\end{table}

%% file: tex/5method.tex
\section{Eyecare Kit: EyecareGPT}

\subsection{Large Vision-Language Models}
The input of LVLMs typically consists of an image $\boldsymbol{x}^\text{img} \in \mathbb{R}^{c\times h \times w}$ and a discrete text sequence $\boldsymbol{x}^\text{txt}$. 
Specifically, a vision encoder $E^\text{img}$ and a text encoder $E^\text{txt}$ are employed to transfer each individual image and the text sequence into a sequence of visual tokens $\mathbf{V} = (v_1, v_2, \ldots, v_{N_\text{img}})$ and textual tokens $\mathbf{T} = (t_1, t_2, \ldots, t_{N_\text{txt}})$, respectively. Subsequently, the visual tokens and the text tokens are fused to form a multi-modal embedding representation $\mathbf{U}=(\mathbf{V}, \mathbf{T})$, which is then fed into a pre-trained large language model $M_\text{LLM}(\cdot\,|\,\theta_\text{LLM})$ for conditional generation. The joint probability of the output response sequence $\mathbf{R} = (r_1, r_2, \ldots, r_{N_\text{res}})$ is modeled as:
\begin{equation}
P_{\theta_\text{LLM}}(\mathbf{R} \mid \mathbf{U}) = \prod_{i=1}^{N_\text{res}} P_{\theta_\text{LLM}}(r_i \mid \mathbf{U}, r_{<i})\, ,
\end{equation}
where $r_i$ is conditioned on the multi-modal input embedding $\mathbf{U}$ and the previously generated tokens $r_{<i}$.

The optimization objective during training is to minimize the cross-entropy loss between the generated sequence and the reference sequence $R^\ast = (r_1^\ast, r_2^\ast,\ldots, r_{N_\text{res}}^\ast)$, which is formulated as:
\begin{equation}
\mathcal{L}(\mathbf{R}^\ast \mid \mathbf{U}) = -\sum_{i=1}^{N_\text{res}} \log P_{\theta_\text{LLM}}(r_i^\ast \mid \mathbf{U}, r_{<i})\, .
\end{equation}

\begin{figure}[t]
    \centering
    \includegraphics[width=\linewidth]{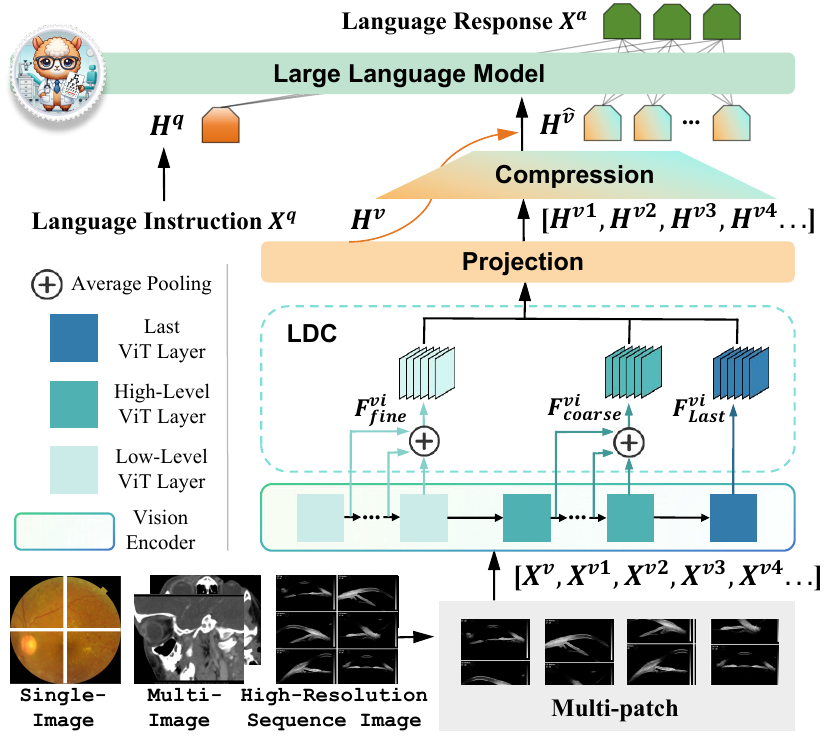}
    \caption{Model architecture of EyecareGPT.}
    \label{fig:model}
    \vskip -0.22in
\end{figure}

\subsection{Layer-wise Dense Connector}

To better preserve fine-grained structural information critical for ophthalmic understanding, we propose a \textbf{Layer-wise Dense Connector} (LDC) that densely aggregates multi-scale visual features from different layers of the vision encoder.

Specifically, given an input image, the vision encoder $M_\text{SLP}(\cdot\,|\,\theta_\text{SLP})$ produces a sequence of hierarchical features $H = (h_1, h_2, \ldots, h_l)$. Conventional approaches typically utilize only the final feature $h_l$, risking the loss of low-level details important for fine-grained tasks. To address this, LDC performs average pooling over shallow-to-intermediate layers and deep layers separately to obtain fine-grained features $F^\text{fine}$ and coarse-grained features $F^\text{coarse}$:
\begin{align}
F^\text{fine}   &= \text{AvgPooling}(h_1, \ldots, h_k)\,, \\
F^\text{coarse} &= \text{AvgPooling}(h_{k+1}, \ldots, h_{l-1})\,.
\end{align}
The final visual representation is obtained by concatenating $F^\text{fine}$, $F^\text{coarse}$, and $h_l$ along the channel dimension:
\begin{equation}
F = \text{Concatenate}(F^\text{fine}, F^\text{coarse}, h_l, \text{dim=channel})\, .
\end{equation}
The fused feature $F$ is then projected to align with the textual embeddings and fed into the LLMs for multi-modal reasoning.

This simple yet effective design enhances fine-grained visual understanding while maintaining overall model efficiency.

\input{table/closed_results}

\input{table/open_results}

\subsection{Adaptive Anyres Mechanism}

Ophthalmic clinical imaging exhibits diverse resolution distributions, complicating consistent visual modeling. To address this, we propose an \textbf{Adaptive Anyres mechanism} that dynamically adapts to varying input scales while preserving fine-grained details.

Specifically, for an input image $x^\text{img}$, if its resolution exceeds a predefined threshold, it is partitioned into four sub-images centered around the center-symmetric point and concatenated with the original image:
\begin{equation}
x^\text{img}_\text{anyres} = (x^\text{img}, x^\text{sub}_1, x^\text{sub}_2, x^\text{sub}_3, x^\text{sub}_4)\, .
\end{equation}
The original and sub-images are encoded by the LDC-based vision encoder $E^\text{LDC}$, and adaptive pooling is applied to the features of the sub-images before concatenating them with the original feature:
\begin{equation}
V = \text{Concat}\left(E^\text{LDC}(x^\text{img}), \, \text{AdaptPooling}\left(E^\text{LDC}(x^\text{sub}_i)_{i=1}^4\right)\right)\,.
\end{equation}

This simple mechanism enables the model to dynamically adapt to heterogeneous resolution distributions, improving the consistency and robustness of multi-modal ophthalmic modeling with minimal computational overhead.

%% file: table/closed_results.tex
\begin{table*}[t]
\centering
\caption{Performance comparison between EyecareGPT-3.8B and other baseline methods on the \textit{closed QA} task from the open-sourced OmniMedVQA and our proposed Eyecare-Bench. \textmd{Where FS. denotes Fluorescein Staining Images modality. We use \textbf{bold} text to indicate the best results and {underline} to indicate the second-best results.}}
\vskip -0.12in
\resizebox{0.95\textwidth}{!}{
\begin{tabular}{lcccccccccc}
\midrule[1.5pt]
\multirow{2}{*}{\textbf{Model}} & \multicolumn{7}{c}{\textbf{Eyecare
-Bench}} & \multicolumn{2}{c}{\textbf{OmniMedVQA}} & \multirow{2}{*}{\textbf{Avg.}}\\
\cmidrule(r){2-8} \cmidrule(l){9-10} 
& \textbf{FS.} & \textbf{Slit-Lamp} & \textbf{OCT} & \textbf{Fundus} & \textbf{FA-ICGA} & \textbf{UBM} & \textbf{CT} & \textbf{OCT} & \textbf{Fundus}\\
\hline
\rowcolor{gray!5} \multicolumn{11}{c}{\textit{Generalist Models}} \\
\hline
LLaVA-1.5-7B~\cite{liu2023llava} & 20.43 & 65.22 & 30.52 & 12.58 & 6.84 & 20.26 & 19.01 & 51.70 & 26.40 & 28.22 \\
Qwen2.5-VL-7B~\cite{Qwen2.5-VL} & 31.74 & {75.71} & 57.86 & {44.90} & 75.79 & 68.66 & {74.65} & 68.74 & 68.46 & {62.95} \\
InternVL-2.5-8B~\cite{chen2025expandingperformanceboundariesopensource} & 32.61 & 58.57 & 52.31 & 37.88 & 73.62 & 62.26 & 61.97 & 78.67 & {77.36} & 59.47 \\
mPLUG-Owl3-7B~\cite{ye2024mplugowl3longimagesequenceunderstanding} & 16.09 & 41.43 & 55.75 & 30.07 & 60.10 & 52.45 & 71.83 & 63.56 & 36.66 & 47.55 \\
Yi-VL-6B~\cite{ai2025yiopenfoundationmodels} & 36.52 & 50.00 & 50.07 & 20.40 & 55.26 & 58.64 & 59.86 & 63.84 & 36.12 & 47.86 \\
MiniCPM-V2.6-8B~\cite{yao2024minicpm} & 25.22 & 58.33 & 59.05 & 16.95 & {79.11} & {73.47} & 66.90 & {86.81} & 77.31 & 60.35 \\
Gemma-3-4B~\cite{team2025gemma} & 22.17 & 71.67 & 46.10 & 25.71 & 60.10 & 63.33 & 33.10 & 53.48 & 57.95 & 48.18 \\
Claude-3.5 & 35.22 & 70.97 & {64.07} & 32.28 & 68.28 & 63.11 & 52.82 & 78.96 & 63.07 & 58.75 \\
\hline
\rowcolor{gray!5} \multicolumn{11}{c}{\textit{Medical Models}} \\
\hline
Med-Flamingo-8.3B~\cite{moor2023med} & 34.78 & 34.48 & 33.16 & 19.39 & 40.57 & 40.94 & 31.69 & 26.96 & 29.11 & 32.34 \\
LLaVA-Med-7B~\cite{li2023llava} & 12.61 & 26.67 & 37.25 & 12.98 & 39.73 & 31.98 & 20.42 & 26.81 & 29.38 & 26.54 \\
MedVLM-R1-2B~\cite{pan2025medvlm} & 31.14 & 64.41 & 59.13 & 42.52 & 55.02 & 56.72 & 63.83 & 71.17 & 76.76 & 57.86 \\
HealthGPT-M3-3.8B~\cite{lin2025healthgpt} & {41.30} & 63.33 & 63.28 & 20.66 & 77.80 & 61.19 & 69.72 & 75.11 & 63.86 & 59.58 \\
\rowcolor{blue!5} \textbf{EyecareGPT-3.8B} & 
\underline{60.87} & \underline{77.03} & \underline{89.76} & \underline{75.10} & \underline{91.43} & \underline{81.66} & \textbf{85.21} & \textbf{100.00} & \textbf{100.00}& \underline{84.56} \\
\rowcolor{blue!5} \textbf{EyecareGPT-7B} & 
\textbf{61.43} & \textbf{77.64} & \textbf{90.09} & \textbf{82.25} & \textbf{92.96} & \textbf{86.78} & \underline{84.33} & \underline{99.26} & \underline{99.56}& \textbf{86.03} \\
\midrule[1.5pt]
\end{tabular}
}
\vskip -0.12in
\label{tab:closed_results}
\end{table*}

%% file: table/open_results.tex
\begin{table*}[t]
\centering
\caption{Performance comparison between EyecareGPT-3.8B and other baseline methods on the \textit{open QA} task from our proposed Eyecare-Bench. \textmd{We use \textbf{bold} text to indicate the best results and {underline} to indicate the second-best results.}}
\vskip -0.12in
\resizebox{0.95\textwidth}{!}{
\begin{tabular}{lcccccccccc}
\midrule[1.5pt]
\multirow{2}{*}{\textbf{Model}} & \multicolumn{2}{c}{\textbf{OCT}} & \multicolumn{2}{c}{\textbf{Fundus}} & \multicolumn{2}{c}{\textbf{FA-ICGA}} & \multicolumn{2}{c}{\textbf{UBM}} & \multicolumn{2}{c}{\textbf{CT}} \\
\cmidrule(r){2-3} \cmidrule(l){4-5} \cmidrule(l){6-7} \cmidrule(l){8-9} \cmidrule(l){10-11}
& \textbf{F1-Bio} & \textbf{Rouge-L} & \textbf{F1-Bio} & \textbf{Rouge-L} & \textbf{F1-Bio} & \textbf{Rouge-L} & \textbf{F1-Bio} & \textbf{Rouge-L} & \textbf{F1-Bio} & \textbf{Rouge-L} \\
\hline
\rowcolor{gray!5} \multicolumn{11}{c}{\textit{Generalist Models}} \\
\hline
LLaVA-1.5-7B~\cite{liu2023llava}            & 52.60 & 15.35 & 67.57 & 18.38 & 22.57 & 7.51  & 62.29 & 15.90 & 27.88 & 10.18\\
Qwen2.5-VL-7B~\cite{Qwen2.5-VL}           & {66.55} & 23.21 & 81.63 & 28.82 & {62.29} & {21.28} & {81.74} & 16.74 & {69.20} & 18.22\\
InternVL-2.5-8B~\cite{chen2025expandingperformanceboundariesopensource}         & 63.51 & 20.94 & 71.44 & 22.92 & 46.84 & 15.86 & 64.24 & 21.03 & 48.78 & 16.68\\
mPLUG-Owl3-7B~\cite{ye2024mplugowl3longimagesequenceunderstanding}           & 42.19 & 19.86 & 79.27 & 25.99 & 31.08 & 9.69  & 56.90 & 22.34 & 60.36 & 18.03\\
Yi-VL-6B~\cite{ai2025yiopenfoundationmodels}                & 56.71 & 20.60 & 71.15 & 22.24 & 17.77 & 7.90  & 59.86 & 20.46 & 31.52 & 14.44\\
MiniCPM-V2.6-8B~\cite{yao2024minicpm}         & 63.60 & {26.88} & 78.13 & 26.92 & 42.71 & 11.95 & 69.20 & 24.10 & 62.83 & 18.88\\
Gemma-3-4B~\cite{team2025gemma}              & 60.29 & 20.45 & 74.48 & 24.93 & 38.67 & 9.20  & 80.96 & 23.83 & 64.04 & {23.16}\\
Claude-3.5              & 62.96 & 21.20 & {85.93} & {28.87} & 42.06 & 12.57 & 78.75 & {26.48} & 62.22 & 16.75\\
\hline
\rowcolor{gray!5} \multicolumn{11}{c}{\textit{Medical Models}} \\
\hline
Med-Flamingo-8.3B~\cite{moor2023med}       & 29.13 & 11.47 & 45.96 & 14.45 & 32.31 & 10.76 & 34.90 & 10.65 & 38.30 & 10.59\\
LLaVA-Med-7B~\cite{li2023llava}            & 51.79 & 23.25 & 82.38 & 26.03 & 32.35 & 9.31  & 68.92 & 20.87 & 66.95 & 17.03\\
MedVLM-R1-2B~\cite{pan2025medvlm}            & 60.29 & 19.26 & 76.46 & 25.10 & 50.40 & 18.13 & 63.53 & 22.72 & 62.83 & 18.42\\
HealthGPT-M3-3.8B~\cite{lin2025healthgpt}       & 51.45 & 13.15 & 61.55 & 16.97 & 56.24 & 17.07 & 71.05 & 21.07 & 57.80 & 15.87\\
\rowcolor{blue!5} \textbf{EyecareGPT-3.8B} 
& \underline{95.53} & \underline{49.68} & \underline{90.79} & \underline{37.03} & \underline{86.75} & \underline{49.21} & \underline{95.78} & \textbf{47.61} & \underline{83.90} & \underline{36.11}\\
\rowcolor{blue!5} \textbf{EyecareGPT-7B} 
& \textbf{96.26} & \textbf{50.10} & \textbf{90.88} & \textbf{38.13} & \textbf{87.86} & \textbf{51.24} & \textbf{96.60} & \underline{47.26} & \textbf{87.27} & \textbf{36.70}\\
\midrule[1.5pt]
\end{tabular}
}
\vskip -0.12in
\label{tab:open_results}
\end{table*}

%% file: tex/6experiments.tex
\input{table/report_results}


\section{Experiments}
\subsection{Data and Experimental Setup}
\noindent \textbf{Data Details.} We follow a three-stage training paradigm, first performing alignment and supervised fine-tuning on a mixture of LLaVA~\cite{liu2023llava} and PubMedVision~\cite{chen2024huatuogpt} data, and then continuing training on Eyecare-100K to enhance domain-specific performance. Detailed settings are provided in the Appendix. We systematically evaluate our model on both the proposed Eyecare-Bench and the existing ophthalmology-related benchmark, OmniMedVQA\cite{hu2024omnimedvqa}, ensuring a comprehensive assessment of its generalization ability and diagnostic performance.

\noindent \textbf{Model Details.} We conduct a zero-shot evaluation on 12 representative LVLMs, including eight open-world LVLMs (e.g., LLaVA-v1.5~\cite{liu2023llava}, Qwen2.5-VL~\cite{Qwen2.5-VL}, InternVL2.5~\cite{chen2025expandingperformanceboundariesopensource}, mPLUG-Owl3~\cite{ye2024mplugowl3longimagesequenceunderstanding}, Yi-VL~\cite{ai2025yiopenfoundationmodels}, MiniCPM-V2.6~\cite{yao2024minicpm}, gemma-3~\cite{team2025gemma}, Claude3.5) and four Med-LVLMs (e.g., Med-Flamingo~\cite{moor2023med}, LLaVA-Med~\cite{li2023llava}, MedVLM-R1~\cite{pan2025medvlm}, HealthGPT~\cite{lin2025healthgpt}). Models (LLaVA-v1.5, mPLUG-Owl3, gemma-3, Med-Flamingo) are not included in the evaluation when they fail to respond to the report generation instructions correctly.

\subsection{Main Experiments}
\noindent \textbf{Visual Question Answering.}
The experimental results for closed and open VQA tasks are presented in Table~\ref{tab:closed_results} and Table~\ref{tab:open_results}, respectively. The main observations are as follows:
\textbf{(i) Superior Performance:} EyecareGPT-3.8B achieves SoTA performance across all evaluation metrics with a relatively small parameter size. For closed VQA, it attains an average accuracy of 84.56\%, significantly outperforming the second-best model at 62.95\% (Qwen2.5-VL-7B). In open VQA, EyecareGPT-3.8B achieves the best F1-Bio score of 90.55, demonstrating outstanding fine-grained ophthalmic image understanding. EyecareGPT-7B, further scaled from a different base model, shows continued performance gains, validating the scalability of the approach.
\textbf{(ii) Limitations of Existing Medical LVLMs:} Due to the lack of high-quality, domain-specific data, existing Med-LVLMs show no significant advantage over general LVLMs in ophthalmic diagnostic tasks. The consistent performance of the EyecareGPT models further highlights the critical role of Eyecare-100K in enhancing fine-grained domain-specific visual understanding.
\textbf{(iii) Multidimensional Evaluation Metrics:} Compared to the existing benchmark OmniMedVQA, Eyecare-Bench covers a broader range of imaging modalities and task types, posing greater challenges and practical evaluation. 
The results demonstrate that Eyecare-Bench effectively reveals performance bottlenecks and deficiencies of LVLMs in ophthalmology, offering valuable insights for model optimization.

\begin{figure}[t]
    \centering
    \includegraphics[width=0.98\linewidth]{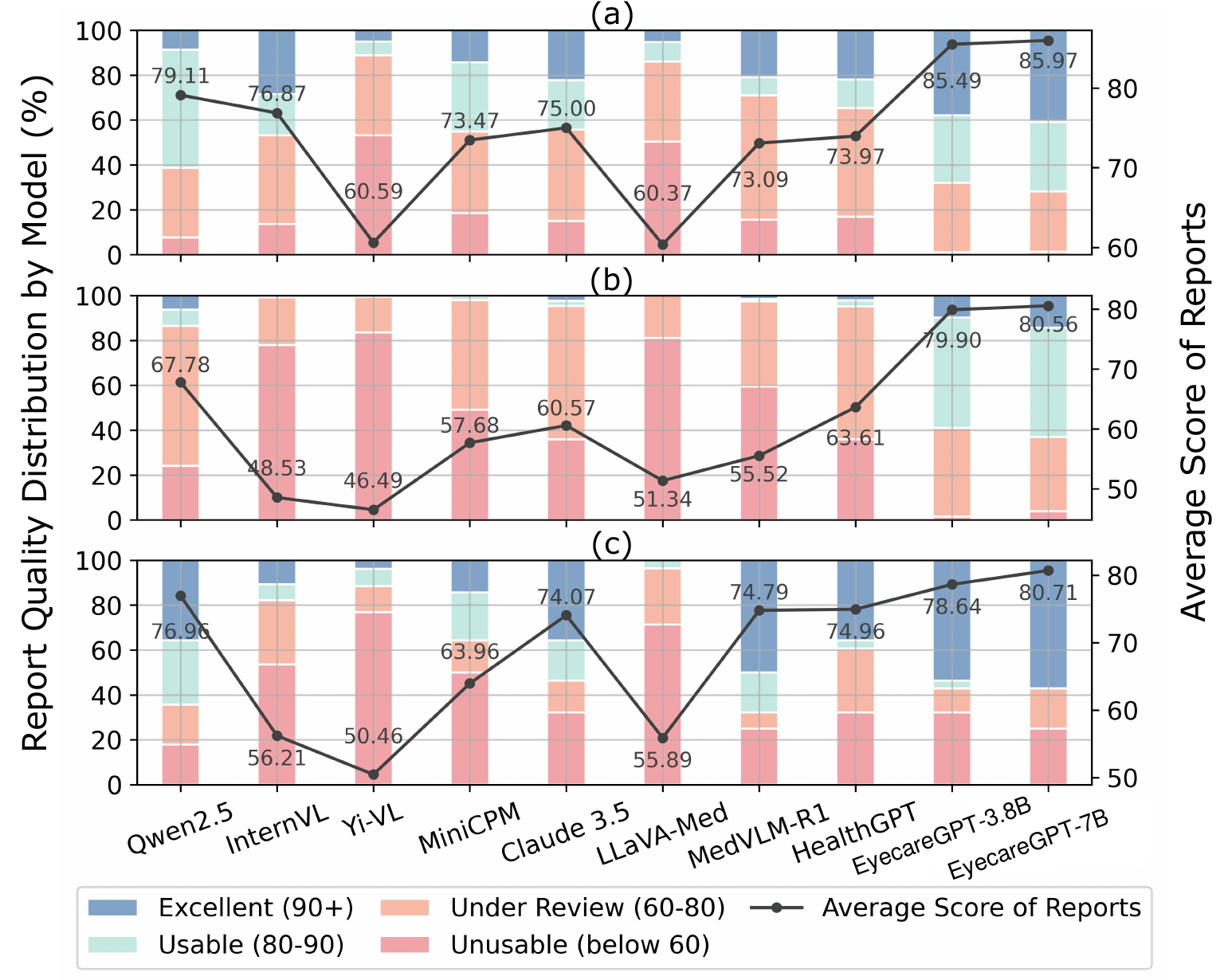}
    \vskip -0.12in
    \caption{GPT-4-based evaluationon results \textmd{for report generation task in (a) FA, (b) UBM, and (c) CT modalities.}}
    \label{fig:data2}
    \vspace{-5mm}
\end{figure}

\noindent \textbf{Report Generation.}
Table~\ref{tab:report_results} presents the experimental results for the report generation task: \textbf{(i)} The EyecareGPT achieves the best performance across all evaluation metrics. Under both GPT-based and traditional evaluations, the diagnostic reports generated by EyecareGPT exceed 50\% accuracy across three imaging modalities and show excellent performance on structural and linguistic consistency metrics such as F1-RadGraph and ROUGE-L. These results demonstrate that EyecareGPT accurately understands complex ophthalmic images and generates high-quality professional reports. \textbf{(ii)} Although some general models (e.g., Qwen2.5, Claude3.5) and medical models (e.g., MedVLM, MedVLM-R1) perform reasonably well in open VQA tasks, they show significant deficiencies in structured medical report generation, failing to maintain scores within a usable range. This highlights the current limitations of LVLMs in handling multi-modal heterogeneous ophthalmic data and the urgent need for targeted optimization.

To further systematically reveal performance differences among models, we introduce a GPT-4-based multi-dimensional evaluation approach, with results shown in Figure~\ref{fig:data2}.
EyecareGPT consistently demonstrates stable and superior performance across all imaging modalities, with more than 50\% of its generated reports rated as clinically usable, showing particularly strong potential in FA and UBM modalities.
In contrast, other models perform relatively better on CT tasks but show clear declines in FA and UBM, reflecting the imbalance of modality distribution in current public datasets. Overall, these results further validate the robustness of EyecareGPT in multi-modal tasks and demonstrate the scientific value of our proposed multi-dimensional evaluation framework in guiding future LVLM optimization.

\subsection{Ablation and In-Depth Study}
\input{table/module_ablation}

\noindent \textbf{Effect of LDC and Anyres.} 
We integrate the Layer-wise Dense Connector (LDC) and the Adaptive Anyres Mechanism to enhance the ability of the model to capture fine-grained structural information and multi-scale visual features, and validate the contribution of each module through ablation studies. As shown in Table~\ref{tab:module_ablation}, LDC, as a plug-and-play module, consistently improves model performance across all three task types by integrating multi-scale visual features. Similarly, the Adaptive Anyres Mechanism strengthens the model’s capability for fine-grained region recognition, achieving notable gains over the baseline. Furthermore, the combined use of LDC and Anyres synergistically balances multi-scale visual modeling and resolution adaptability, significantly boosting diagnostic accuracy and report generation quality, thereby demonstrating the effectiveness and generalizability of the architecture in intelligent ophthalmic diagnosis.

\noindent \textbf{Ablation on Eyecare-100K.}
\begin{figure}[t]
    \centering
    \includegraphics[width=0.95\linewidth]{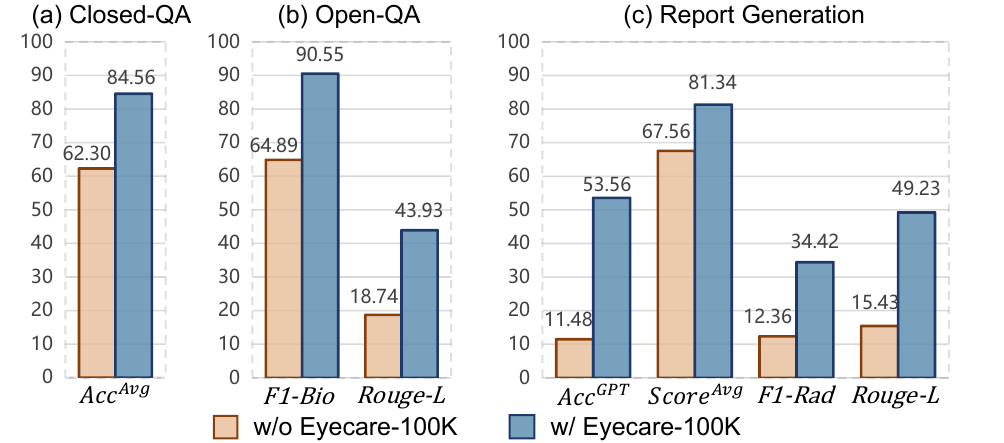}
    \vskip -0.12in
    \caption{Results after fine-tuning on Eyecare-100K.}
    \label{fig:dataablation}
    \vskip -0.12in
\end{figure}
We validate Eyecare-100K's effectiveness in enhancing visual understanding of complex ophthalmic scenarios by comparing model performance before and after fine-tuning, as shown in Figure~\ref{fig:dataablation}. 
In closed VQA tasks, the average accuracy of the model improves from 65.30\% to 84.56\%, reaching 100\% accuracy on the OmniMedVQA subtask. In open VQA and report generation tasks, the average F1-Bio score increases from 64.89 to 90.55, and the GPT-evaluated diagnostic accuracy rises from 11.48\% to 53.56\%. These significant improvements demonstrate the high quality and broad applicability of Eyecare-100K in constructing multi-modal, multi-task instruction datasets.

\begin{figure}[t]
    \centering
    \includegraphics[width=\linewidth]{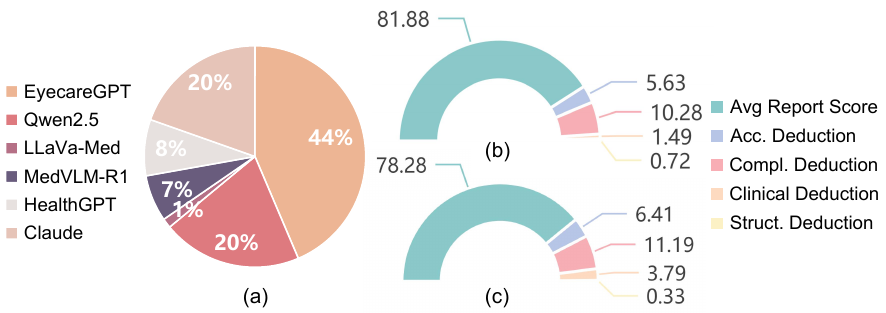}
    \vskip -0.12in
    \caption{ Physician preference \textmd{for generated reports (a) and EyeEval reliability (b, c).}}
    \label{fig:human}
    \vskip -0.12in
\end{figure}

\noindent \textbf{Expert Physician Review.}
We further conduct an expert physician review of the report generation task in Eyecare-Bench. We recruited ten clinicians to rank the responses from EyecareGPT-3.8B, Qwen-VL-7B, Claude 3.5, LLaVA-Med, MedVLM-R1, and HealthGPT-M3, and to select the report with the highest satisfaction. We randomly sample 500 report generation VQA pairs along with the answers generated by the aforementioned six models and randomly order them for the doctors to choose from. The final results of the doctors' selections are shown in Figure~\ref{fig:human} (a), indicating that the reports generated by EyecareGPT are more satisfactory to clinicians and better meet clinical needs. Simultaneously, we also asked the doctors to score the reports generated by EyecareGPT according to our proposed EyeEval evaluation system. The report scores and deduction details across four dimensions evaluated by GPT-4 and doctors are shown in Figure~\ref{fig:human} (b) and (c) respectively, demonstrating that EyeEval and doctor diagnoses have high consistency and reliability.

%% file: table/report_results.tex
\begin{table*}[t]
\centering
\caption{Performance comparison between EyecareGPT-3.8B and other baseline methods on the \textit{report generation} task from our proposed Eyecare-100K benchmark. \textmd{We use \textbf{bold} text to indicate the best results and \underline{underline} to indicate the second-best results.}}
\vskip -0.12in
\resizebox{0.93\textwidth}{!}{
\begin{tabular}{lccccccccc}
\midrule[1.5pt]
\multirow{2}{*}{\textbf{Model}} & \multicolumn{3}{c}{\textbf{FA-ICGA}} & \multicolumn{3}{c}{\textbf{UBM}} & \multicolumn{3}{c}{\textbf{CT}} \\
\cmidrule(r){2-4} \cmidrule(l){5-7} \cmidrule(l){8-10} 
& \textbf{Acc\textsuperscript{GPT}} & \textbf{F1-Rad} & \textbf{Rouge-L} & \textbf{Acc\textsuperscript{GPT}}  & \textbf{F1-Rad} & \textbf{Rouge-L} & \textbf{Acc\textsuperscript{GPT}} & \textbf{F1-Rad} & \textbf{Rouge-L} \\
\hline
\rowcolor{gray!5} \multicolumn{10}{c}{\textit{Generalist Models}} \\
\hline
Qwen2.5-VL-7B~\cite{Qwen2.5-VL}           & 17.00 & 6.91 & 15.54 & 19.54 & 4.26 & 7.21 & 42.86 & 11.37 & 18.23\\
InternVL-2.5-8B~\cite{chen2025expandingperformanceboundariesopensource}         & 5.92 & 5.19 & 8.51  & 0.00 & 3.53 & 8.39 & 0.00 & 8.00 & 12.68 \\
Yi-VL-6B~\cite{ai2025yiopenfoundationmodels}                & 2.26 & 5.12 & 9.13  & 0.00 & 2.14 & 8.03  & 3.85 & 6.73 & 14.68\\
MiniCPM-V2.6-8B~\cite{yao2024minicpm}         & 3.34 & 6.12 & 8.56 & 0.00 & 3.78 & 7.33 & 0.00 & 5.01 & 11.88\\
Claude-3.5              & 14.53 & 6.37 & 12.96 & 2.98 & 9.56 & 14.38 & 25.00 & 11.05 & 16.23\\
\hline
\rowcolor{gray!5} \multicolumn{10}{c}{\textit{Medical Models}} \\
\hline
LLaVA-Med-7B~\cite{li2023llava}            & 0.14 & 3.53 & 12.64 & 0.00 & 4.63 &  8.32 & 0.00 & 1.02 & 12.15\\
MedVLM-R1-2B~\cite{pan2025medvlm}            & 11.28 & 5.76 & 4.38 & 2.22 & 4.05 & 6.41 & 32.14 & 11.87 & 10.64\\
HealthGPT-M3-3.8B~\cite{lin2025healthgpt}       & 14.41 & 7.30 & 12.86 & 3.02 & 5.53 & 10.30 & 17.35 & 14.47 & 18.21\\
\rowcolor{blue!5} \textbf{EyecareGPT-3.8B} 
& \underline{52.62} & \underline{25.04} & \underline{47.91} & \underline{58.05} & \underline{42.83} & \underline{57.04} & \underline{50.00} & \underline{35.39} & \underline{42.73}\\
\rowcolor{blue!5} \textbf{EyecareGPT-7B} 
& \textbf{53.91} & \textbf{26.04} & \textbf{48.32} & \textbf{60.06} & \textbf{42.98} & \textbf{58.43} & \textbf{52.43} & \textbf{36.19} & \textbf{43.54}\\
\midrule[1.5pt]
\end{tabular}
}
\vskip -0.12in
\label{tab:report_results}
\end{table*}

%% file: table/module_ablation.tex
\begin{table*}[t]
\centering
\caption{Ablation study of the effect of the individual module for three tasks.}
\vskip -0.12in

\resizebox{0.98\textwidth}{!}{
\begin{tabular}{c|cc|cccccccccc}
\midrule[1.5pt]
\multirow{2}{*}{\textbf{Task}} & \multirow{2}{*}{\textbf{AnyRes}} &\multirow{2}{*}{\textbf{LDC}} & \multicolumn{7}{c}{\textbf{Eyecare-100K}} & \multicolumn{2}{c}{\textbf{OmniMedVQA}} & \multirow{2}{*}{\textbf{Avg.}}\\
\cmidrule(r){4-10} \cmidrule(l){11-12} 
 & & & \textbf{FS.} & \textbf{SL.} & \textbf{OCT} & \textbf{Fundus} & \textbf{FA-ICGA} & \textbf{UBM} & \textbf{CT} & \textbf{OCT} & \textbf{Fundus}\\
\hline
\multirow{4}{*}{\rotatebox[origin=c]{90}{Closed-QA}}& - & - & 60.00 & 68.92 & 87.19 & 73.77 & 88.62 & 78.25 & 82.39 & 99.26 & 98.65 &  81.89\\
& \faCheckCircle & - & \underline{60.71} & \underline{76.47} & \underline{88.61} & \textbf{76.42} & \underline{90.78} & \underline{80.55} & \underline{83.95} & 99.41 & \textbf{100.00} & \underline{84.10} \\
 & - & \faCheckCircle & 60.00 & 70.23 & 87.58 & 74.78 & 89.20 & 79.96 & 83.45 & \underline{99.50} & \underline{99.56} & 82.79 \\
 & \cellcolor{blue!5}\faCheckCircle & \cellcolor{blue!5}\faCheckCircle  & \cellcolor{blue!5}\textbf{60.87} & \cellcolor{blue!5}\textbf{77.03} & \cellcolor{blue!5}\textbf{89.76} & \cellcolor{blue!5}75.10 & \cellcolor{blue!5}\textbf{91.43} & \cellcolor{blue!5}\textbf{81.66} & \cellcolor{blue!5}\textbf{85.21} & \cellcolor{blue!5}\textbf{100.00} & \cellcolor{blue!5}\textbf{100.00}& \cellcolor{blue!5}\textbf{84.56} \\
\midrule[1pt]
\multirow{2}{*}{\textbf{Task}} & \multirow{2}{*}{\textbf{AnyRes}} &\multirow{2}{*}{\textbf{LDC}} &\multicolumn{2}{c}{\textbf{OCT}} & \multicolumn{2}{c}{\textbf{Fundus}} & \multicolumn{2}{c}{\textbf{FA-ICGA}} & \multicolumn{2}{c}{\textbf{UBM}} & \multicolumn{2}{c}{\textbf{CT}} \\
\cmidrule(r){4-5} \cmidrule(r){6-7} \cmidrule(r){8-9} \cmidrule(r){10-11} \cmidrule(r){12-13} 
& & & \textbf{F1-Bio} & \textbf{Rouge-L} & \textbf{F1-Bio} & \textbf{Rouge-L} & \textbf{F1-Bio} & \textbf{Rouge-L} & \textbf{F1-Bio} & \textbf{Rouge-L} & \textbf{F1-Bio} & \textbf{Rouge-L} \\
\hline
\multirow{4}{*}{\rotatebox[origin=c]{90}{Open-QA}}& - & - & 95.12 & 48.32 &  87.15 & 36.46 & 85.89 & 46.77 & 93.48 & 45.25 & 79.38 & 32.23\\
& \faCheckCircle & - 
& \underline{95.38} & \underline{49.17} & \underline{89.76} & \underline{36.90} & \underline{86.34} & \underline{48.15} & \underline{95.66} & \underline{46.94} & \underline{82.23} & \underline{33.43} \\
 & - & \faCheckCircle 
& 95.20 & 48.98 & 88.90 & 36.77 & 85.95 & 47.75 & 94.73 & 45.79 & 80.98 & 32.93 \\
 & \cellcolor{blue!5}\faCheckCircle & \cellcolor{blue!5}\faCheckCircle  & \cellcolor{blue!5}\textbf{95.53} & \cellcolor{blue!5}\textbf{49.68} & \cellcolor{blue!5}\textbf{90.79} & \cellcolor{blue!5}\textbf{37.03} & \cellcolor{blue!5}\textbf{86.75} & \cellcolor{blue!5}\textbf{49.21} & \cellcolor{blue!5}\textbf{95.78} & \cellcolor{blue!5}\textbf{47.61} & \cellcolor{blue!5}\textbf{83.90} & \cellcolor{blue!5}\textbf{36.11} \\
 \midrule[1pt]
 \end{tabular}}

 \resizebox{0.98\textwidth}{!}{
 \begin{tabular}{c|cc|cccccccccccc}
 \multirow{2}{*}{\textbf{Task}} & \multirow{2}{*}{\textbf{AnyRes}} &\multirow{2}{*}{\textbf{LDC}} & \multicolumn{4}{c}{\textbf{FA-ICGA}} & \multicolumn{4}{c}{\textbf{UBM}} & \multicolumn{4}{c}{\textbf{CT}} \\
\cmidrule(r){4-7} \cmidrule(r){8-11} \cmidrule(r){12-15} 
&  & & \textbf{Acc\textsuperscript{GPT}} &  \textbf{Score\textsuperscript{Avg}}& \textbf{F1-Rad} & \textbf{Rouge-L} & \textbf{Acc\textsuperscript{GPT}}  &  \textbf{Score\textsuperscript{Avg}} & \textbf{F1-Rad} & \textbf{Rouge-L} & \textbf{Acc\textsuperscript{GPT}} &  \textbf{Score\textsuperscript{Avg}} & \textbf{F1-Rad} & \textbf{Rouge-L} \\
\hline
\multirow{4}{*}{\rotatebox[origin=c]{90}{Report-Gen}}& - & - 
& 45.37 & 83.02 & 23.56 & 46.71 & 32.48 & 72.36 & 39.14 & 53.78 & 42.71 & 76.71 & 29.42 & 40.47\\
& \faCheckCircle & - 
& \underline{51.37} & \underline{85.13} & \underline{24.76} & \underline{47.80} & \underline{54.86} & \underline{79.07} & \underline{41.36} & \underline{56.10} & \underline{45.88} & \underline{78.35} & \underline{32.27} & \underline{41.52} \\
 & - & \faCheckCircle 
& 47.83 & 84.85 & 24.62 & 46.81 & 44.42 & 77.45 & 40.29 & 54.43 & 44.97 & 77.25 & 30.33 & 40.65 \\
 & \cellcolor{blue!5}\faCheckCircle & \cellcolor{blue!5}\faCheckCircle 
& \cellcolor{blue!5}\textbf{52.62} & \cellcolor{blue!5}\textbf{85.49} & \cellcolor{blue!5}\textbf{25.04} & \cellcolor{blue!5}\textbf{47.91} & \cellcolor{blue!5}\textbf{58.05} & \cellcolor{blue!5}\textbf{79.90} & \cellcolor{blue!5}\textbf{42.83} & \cellcolor{blue!5}\textbf{57.04} & \cellcolor{blue!5}\textbf{50.00} & \cellcolor{blue!5}\textbf{78.64} & \cellcolor{blue!5}\textbf{35.39} & \cellcolor{blue!5}\textbf{42.73} \\
 \midrule[1.5pt]
\end{tabular}
}

\label{tab:module_ablation}
\end{table*}

%% file: tex/7conclusion.tex
\section{Conclusion}

We propose \textbf{Eyecare Kit}, a comprehensive framework that addresses critical challenges in intelligent ophthalmic diagnosis through tailored dataset, benchmark, and model.
Experiments validate the effectiveness of \textbf{Eyecare-100K}, \textbf{Eyecare-Bench}, and \textbf{EyecareGPT}, with EyecareGPT achieving SOTA results. We believe Eyecare Kit lays a solid foundation for future advances in domain-specific Med-LVLMs and ophthalmic AI applications.

%% file: tex/Appendix.tex
\newpage
\newpage
\begin{center}
\section*{Appendix}
\end{center}

This is the Appendix for ``EyecareGPT: Boosting Comprehensive Ophthalmology
Understanding with Tailored Dataset, Benchmark and Model''. This Appendix is organized as follows:

\begin{itemize}
\item \textbf{Section A} presents the details of the experimental implementation, the training process of \textbf{EyecareGPT}, the construction details of \textbf{Eyecare-100K}, and the specific information of \textbf{Eyecare-Bench}.
\item \textbf{Section B} presents our more detailed ablation experimental results and a brief experimental analysis.
\item \textbf{Section C} shows typical data examples in \textbf{Eyecare-100K}.
\end{itemize}

\section{Implementation Details}
\subsection{Model Details}

We use SigLIP-SO400M-Patch14-384 as the visual feature extractor to capture multi-scale visual features, ensuring the model’s ability to recognize and understand local details. The visual features are aligned with text embeddings through an MLP and jointly fed into the large language model for conditional output.

EyecareGPT offers two versions: \textbf{EyecareGPT-3.8B} and \textbf{EyecareGPT-7B}, which are based on Phi-3.5-mini-Instruct and Qwen2.5-7B-Instruct as the pre-trained LLMs, respectively. Table \ref{tab:model_details} shows the details.
\begin{table}[h!]
    \centering
    \vspace{-1mm}
    \caption{Overview of the components of EyecareGPT.}
    \resizebox{0.8\textwidth}{!}{
    \begin{tabular}{l|ccccc}
        \midrule[1.5pt]

        \textbf{Model} & \textbf{ViT} & \textbf{Adapter}  & \textbf{LLM} & \textbf{Params}& \textbf{LoRA Rank} \\ \hline \hline
\rowcolor{blue!5}
        EyecareGPT-3.8B & SigLIP-SO400M & 2-layer MLP  & Phi-3.5-mini-Instruct & 3.8B  & 64\\
\rowcolor{blue!5}
        EyecareGPT-7B & SigLIP-SO400M & 2-layer MLP & Qwen2.5-7B-Instruct & 7B  & 64 \\ 
        \midrule[1.5pt]
    \end{tabular}
    }
    \label{tab:model_details}
\end{table}
\vspace{-2mm}
\subsection{Training Details}
This study adopts a three-stage training strategy to progressively build the model’s visual understanding and intelligent ophthalmic diagnostic capabilities. In the first stage, we train the model on data for aligning from LLaVA-558K and PubMedVision to enhance image description and basic vision-language alignment capabilities. In the second stage, we use supervised fine-tuning data from LLaVA-665K and PubMedVision to further strengthen the model’s visual instruction following and general medical understanding. In the third stage, we perform specialized fine-tuning on Eyecare-100K, focusing on three core task types in clinical ophthalmic diagnosis while avoiding noise introduced by other data sources, thereby optimizing the model’s domain-specific adaptability and accuracy. Hyperparameter configurations for each training stage are detailed in Table~\ref{tab:parameters} to ensure training efficiency and convergence.

\begin{table}[h!]
    \centering
    \caption{Overview of hyperparameter configurations.}
    \resizebox{0.75\textwidth}{!}{
    \begin{tabular}{l|c|c|c|c|c|c}
    \midrule[1.5pt]
    \rowcolor{blue!5}
    & \multicolumn{3}{c|}{\textbf{EyecareGPT-3.8B}} & \multicolumn{3}{c}{\textbf{EyecareGPT-7B}} \\
    \cline{2-7}
    \rowcolor{blue!5}
    \multirow{-2}{*}{\textbf{Hyperparameter}} & \textbf{Stage-1} & \textbf{Stage-2} & \textbf{Stage-3} & \textbf{Stage-1} & \textbf{Stage-2} & \textbf{Stage-3} \\
    \hline \hline
    Optimizer & AdamW & AdamW & AdamW & AdamW & AdamW & AdamW \\
    Adapter LR & 1e-3 & 2e-5 & 2e-5 & 1e-3 & 2e-5 & 2e-5 \\
    Learning Rate & / & 2e-4 & 2e-4 & / & 2e-4 & 2e-4 \\
    Global Batch Size & 256 & 128 & 32 & 256 & 128 & 32 \\
    Weight Decay & 0 & 0 & 0 & 0 & 0 & 0 \\
    Dropout Rate & 0 & 0.05 & 0.05 & 0 & 0.05 & 0.05 \\
    LR Scheduler & Warm Up & Warm Up & Constant & Warm Up & Warm Up & Constant \\
    Max Sequence Length & 2048 & 2048 & 2048 & 2048 & 2048 & 2048 \\ 
    \midrule[1.5pt]
    \end{tabular}
    }
    \label{tab:parameters}
\end{table}

\subsection{Construction details of Eyecare-100K}
\noindent \textbf{Data Source Details:} 
In the data collection phase, we gathered eye report data with four modalities – Fluorescein Angiography (FA), Indocyanine Green Angiography (ICGA), Ultrasound Biomicroscopy (UBM), and Computed Tomography (CT) – from our hospital. Specifically, this included 2081 CT images, 3425 UBM images, 15048 FA images, and 2394 ICGA images. Furthermore, to fully leverage real-world data from existing public datasets, we collected 10 previously published ophthalmological datasets with Fundus and Optical Coherence Tomography (OCT) modalities, as detailed in Table~\ref{tab:retinal_datasets}. These are all single-modality datasets containing disease classification or grading labels and corresponding images. We also collected three publicly available datasets from Kaggle, including one fluorescence-stained image dataset, one slit-lamp dataset, and one OCT dataset.

\begin{figure*}[h!]
    \centering
    \includegraphics[width=0.615\linewidth]{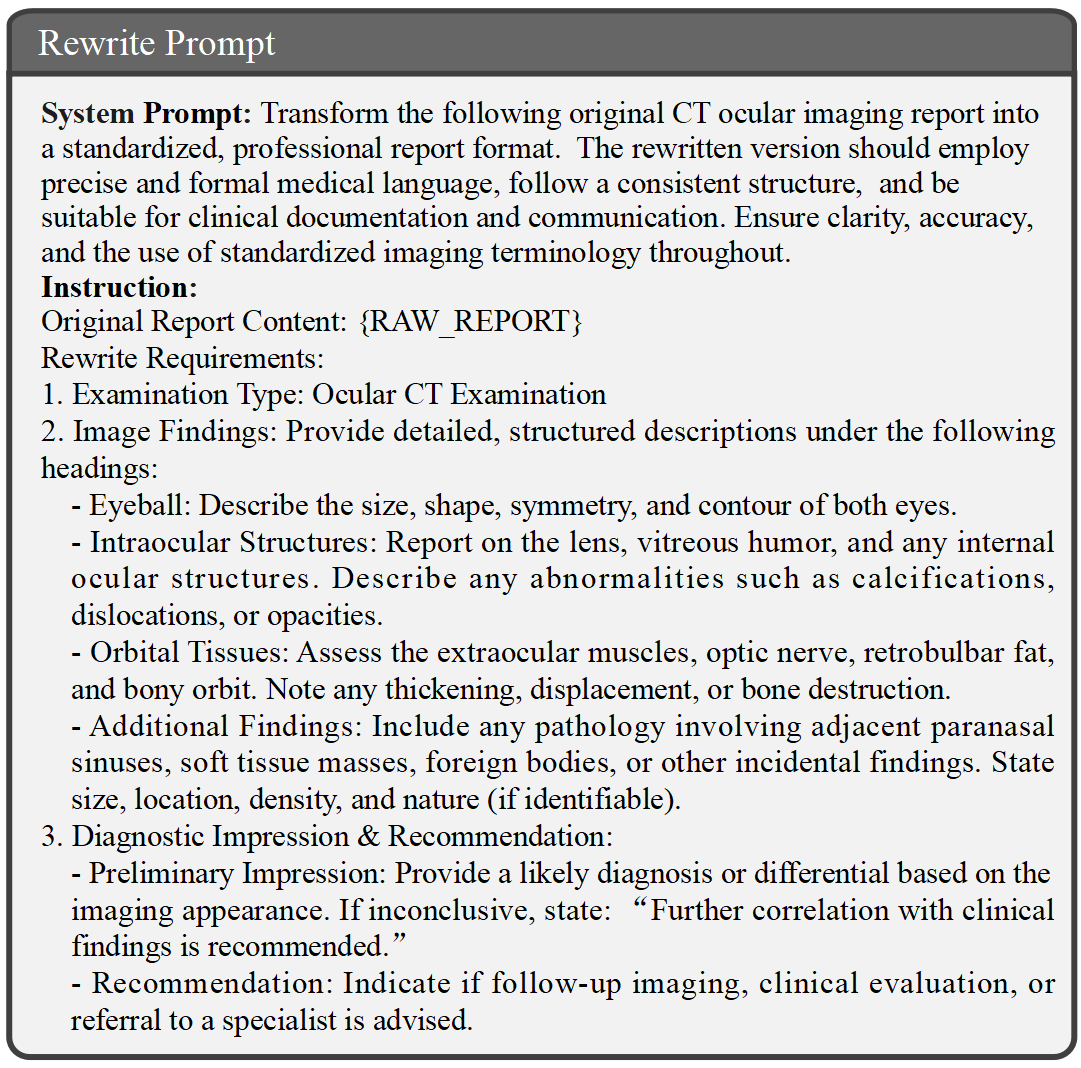}
    \caption{Rewrite Prompt}
    \label{fig:rewrite-Prompt}
\end{figure*}

\begin{table*}[htbp]
\centering
\caption{Overview of Existing Eye Publicly Available Datasets Collected }
\label{tab:retinal_datasets}
\resizebox{0.8\textwidth}{!}{
\begin{tabular}{@{}lllll@{}}
\toprule
\textbf{Dataset Name} & \textbf{Modality Type} & \textbf{Source} \\
\midrule
IDRID & Fundus & Aravind Eye Hospital, Madurai, India \\
ACRIMA& Fundus & University of Jaén, Spain  \\
JSIEC & Fundus & Joint Shantou University - Eye Center, China  \\
ODIR& Fundus & Multiple hospitals in China  \\
MuReD& Fundus & Multiple hospitals in China  \\
DeepDRID & Fundus & Multiple hospitals in India  \\
OCT2017 & OCT & Shiley Eye Institute, University of California San Diego, USA  \\
OCTID& OCT & Zhongshan Ophthalmic Center, Sun Yat-sen University, China  \\
OCTDL & OCT & University of Tuebingen, Germany  \\
Kermany & OCT & Multiple sources (publicly available)  \\
\bottomrule
\end{tabular}  
}
\end{table*}
\noindent \textbf{Rewrite Prompt:} 
For the report data collected from the hospital, we processed it and used Claude 3.7 to construct three types of data. Through discussions with doctors, we identified three key components in the reports: Image Type, Imaging Findings, and Diagnostic Suggestions. Therefore, we designed prompts to guide Claude in generating reports with a unified structure. Figure~\ref{fig:rewrite-Prompt} illustrates the prompt template we designed for the CT modality.

\begin{figure*}[h!]
    \centering
    \includegraphics[width=0.61\linewidth]{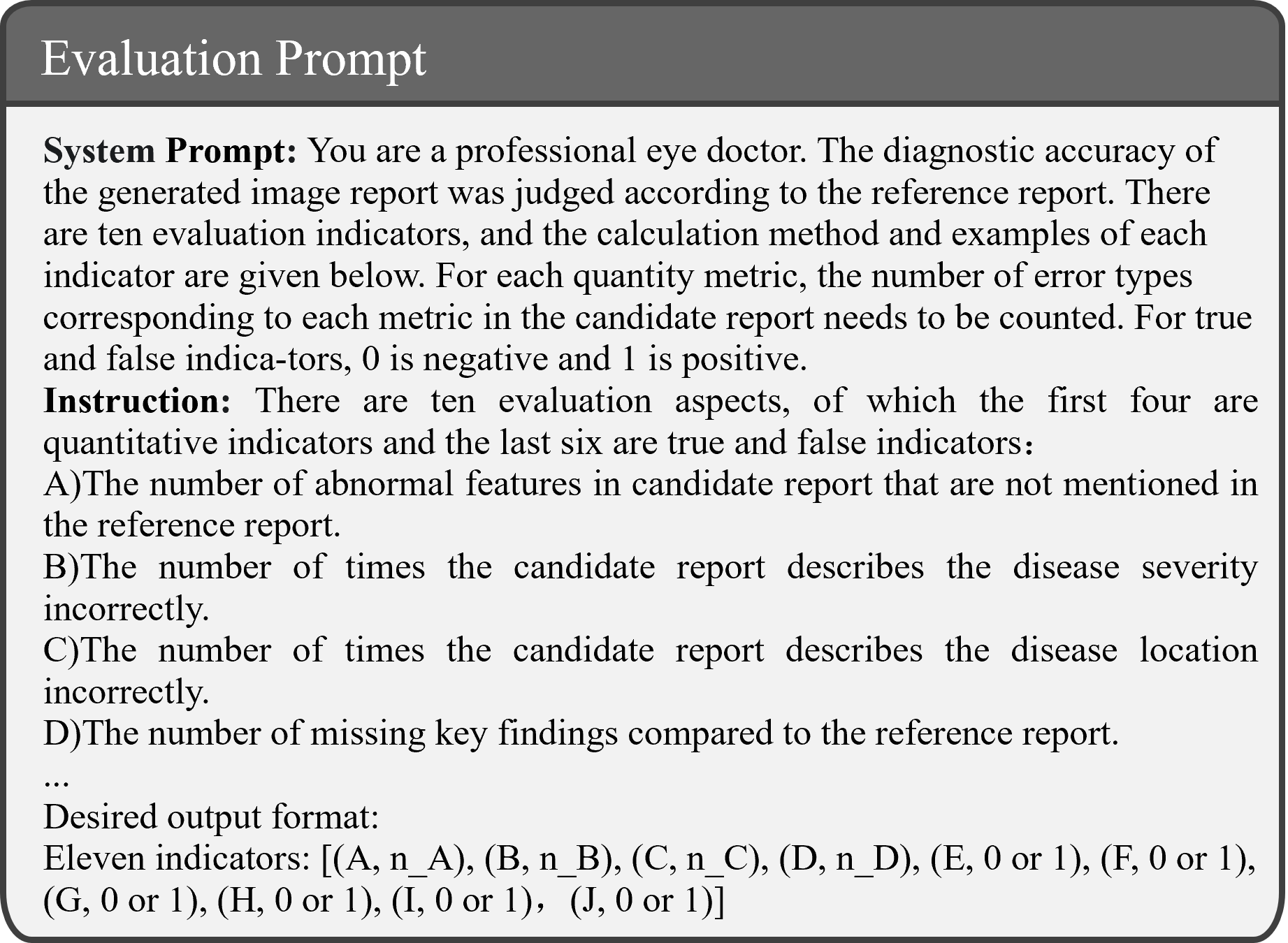}
    \caption{Evaluation Prompt}
    \label{fig:evaluation}
\end{figure*}

\noindent \textbf{QA Templates:} 
For the aforementioned datasets that only contain classification or grading labels, we analyzed the data characteristics of their labels and designed different Question-Answering (QA) templates for each. This allowed us to transform the original data into open-ended Question-Answering pairs. Examples of the QA templates are shown in the Table~\ref{tab:medical_comparison}.

\begin{table*}[htbp]
\centering
\caption{Sample Question Answering (QA) Templates for Data Conversion.}
\resizebox{0.65\textwidth}{!}{
\begin{tabular}{p{15cm}}

\hline
\textbf{Question1:} \\
\begin{enumerate}[label=\arabic*., leftmargin=10pt]
\item Is the eye in this picture diseased?.
\item Does the eye shown in the image have any disease?
\item Is there any sign of illness in the eye in this photo?
\item Does this eye image show any signs of abnormalities?
\item Does the eye in the image show signs of disease?
\item Is there evidence of a disorder in the eye in this picture?
\item Are there any visible abnormalities in the eye image?
\end{enumerate}\\

\hline
\textbf{Positive <condition>:} \\
\begin{enumerate}[label=\arabic*., leftmargin=10pt]
\item Yes, the eye in the picture has \{condition\}.
\item Yes, the image reveals the presence of \{condition\} in the eye.
\item Yes, the eye shown in this image is impacted by \{condition\}.
\item Yes, this image depicts an eye presenting \{condition\}.
\item Yes, the eye in this image shows evidence of \{condition\}.
\item Yes, the image illustrates an eye with \{condition\}.
\end{enumerate}\\

\hline
\textbf{Negative <condition>:} \\
\begin{enumerate}[label=\arabic*., leftmargin=10pt]
\item No, very healthy.
\item No, the eye appears healthy in the image.
\item No. This image shows that the retina looks normal, with no hemorrhages, exudates or other signs of abnormality.
\item No, the eye image appears normal.
\item No, the findings from the retinal image suggest a normal and healthy eye.
\item No, there are no indications of disease in the image.
\item No, the retinal image indicates a healthy eye, with no signs of hemorrhages, exudates, or other pathological changes.
\item No significant abnormalities were detected in the eye image.
\end{enumerate}\\

\hline
\textbf{Question2:} \\
\begin{enumerate}[label=\arabic*., leftmargin=10pt]
\item What ocular disease is evident in this image?
\item What eye condition is visible in this picture?
\item What condition is affecting the eye shown in the image?
\item What issue is apparent in the eye shown here?
\item What is wrong with the eye in the image?
\item Which disease can be seen in the eye from this picture?
\item What health issue is present in the eye in this image?
\item What health concern is evident in the eye in this image?
\item What problem does the eye shown in the image have?
\end{enumerate}\\

\hline
\textbf{Positive <condition>:} \\
\begin{enumerate}[label=\arabic*., leftmargin=10pt]
\item The eye in the image exhibits signs of \{condition\}.
\item \{condition\} is evident in the eye depicted in the image.
\item The image reveals the presence of \{condition\} in the eye.
\item In this picture, the eye appears to be affected by \{condition\}.
\item This image shows an eye with \{condition\}.
\item The eye in the photograph shows signs of \{condition\}.
\item \{condition\} is visible in the eye from this picture.
\end{enumerate}\\
\hline
\textbf{Negative <condition>:} \\
\begin{enumerate}[label=\arabic*., leftmargin=10pt]
\item The eye in this image is very healthy.
\item This picture shows a perfectly healthy eye with no signs of disease.
\item The eye depicted in the image is completely healthy, showing no illness.
\item There is no indication of disease in the eye shown by this image. It's very healthy.
\item According to this image, the eye is very healthy and free from any disease.
\item The photo indicates a very healthy eye with no presence of disease.
\end{enumerate}\\
\hline
\end{tabular}}
\label{tab:medical_comparison}
\end{table*}

\subsection{GPT-4 Evaluation Prompt}
We designed an evaluation system called EyeEval and introduced GPT-4 for the evaluation process. The template used for GPT-4's evaluation is shown in Figure~\ref{fig:evaluation}. According to the scoring criteria, we grade the reports as follows:
\begin{itemize}
    \item \textbf{Excellent Report (90-100):} The report contains virtually no errors, the information is relatively complete, the structure is clear, and it does not contain serious clinical errors.
    \item \textbf{Usable Report (80-90):} The report may contain some minor errors, but overall the information is complete, the structure is clear, and it does not contain serious clinical errors.
    \item \textbf{Report Under Review (60-80):} The report contains numerous errors or missing information, the diagnosis may be inaccurate, or the report structure is disorganized, requiring further review.
    \item \textbf{Unusable Report (Below 60):} The report contains a large number of errors, severely missing information, diagnostic errors, or contains serious clinical errors, making it unsuitable for clinical decision-making.
\end{itemize}

\begin{table*}[htbp]
\centering
\caption{Other evaluation metrics for the open-ended question answering task in the main experiment. }
\resizebox{\textwidth}{!}{
\begin{tabular}{lcccccccccccccccccc}
\midrule[1.5pt]
\multirow{2}{*}{\textbf{Model}} & \multicolumn{3}{c}{\textbf{OCT}} & \multicolumn{3}{c}{\textbf{Fundus}} & \multicolumn{3}{c}{\textbf{FA-ICGA}} & \multicolumn{3}{c}{\textbf{CT}} & \multicolumn{3}{c}{\textbf{UBM}} \\
\cmidrule(r){2-4} \cmidrule(l){5-7} \cmidrule(l){8-10} \cmidrule(l){11-13} \cmidrule(l){14-16}
& \textbf{F1-Rad} & \textbf{BLEU-1} & \textbf{BLEU-4} & \textbf{F1-Rad} & \textbf{BLEU-1} & \textbf{BLEU-4} & \textbf{F1-Rad} & \textbf{BLEU-1} & \textbf{BLEU-4} & \textbf{F1-Rad} & \textbf{BLEU-1} & \textbf{BLEU-4} & \textbf{F1-Rad} & \textbf{BLEU-1} & \textbf{BLEU-4} \\
\hline
\rowcolor{gray!5} \multicolumn{16}{c}{\textit{Generalist Models}} \\
\hline
LLaVA-1.5           & 8.50  & 11.20 & 2.18  & 6.76  & 28.57 & 2.44  & 3.33  & 3.01  & 0.26  & 7.48  & 6.58  & 0.86  & 15.69 & 12.69 & 1.93  \\
Qwen2.5-VL           & 13.39 & 22.23 & 5.06  & 20.46 & 36.45 & 10.21 & 6.12  & 11.12 & 2.44  & 11.37 & 16.28 & 2.95  & 15.91 & 6.13  & 1.15  \\
InternVL-2.5       & 12.90 & 20.06 & 4.43  & 16.75 & 30.09 & 7.30  & 4.38  & 10.49 & 1.38  & 9.39  & 17.02 & 3.32  & 17.75 & 25.34 & 4.50  \\
mPLUG-Owl3          & 10.57 & 16.63 & 3.05  & 21.26 & 30.02 & 7.14  & 4.61  & 6.37  & 6.00  & 10.77 & 15.99 & 3.12  & 19.02 & 20.70 & 3.57  \\
Yi-VL               & 10.71 & 17.02 & 3.24  & 16.43 & 19.68 & 4.31  & 2.06  & 6.92  & 0.59  & 10.89 & 11.33 & 1.77  & 15.43 & 17.75 & 3.05  \\
MiniCPM-V2.6       & 14.92 & 30.48 & 8.78  & 19.51 & 30.76 & 8.42  & 6.17  & 10.18 & 1.70  & 12.79 & 17.21 & 3.44  & 20.52 & 27.64 & 5.26  \\
Gemma-3              & 9.20  & 23.56 & 5.50  & 17.65 & 32.76 & 7.54  & 4.71  & 6.49  & 7.00  & 16.81 & 24.76 & 4.47  & 17.87 & 25.44 & 4.88  \\
\hline
\rowcolor{gray!5} \multicolumn{16}{c}{\textit{Medical Models}} \\
\hline
LLaVA-Med            & 12.36 & 22.74 & 4.58  & 19.44 & 28.09 & 7.03  & 6.45  & 6.77  & 0.73  & 14.34 & 14.51 & 2.52  & 18.89 & 19.04 & 3.48  \\
MedVLM-R1            & 10.08 & 22.06 & 4.30  & 18.82 & 28.68 & 7.18  & 6.34  & 7.73  & 0.77  & 13.83 & 14.43 & 2.94  & 17.51 & 24.08 & 4.31  \\
HealthGPT-M3       & 6.64  & 14.12 & 2.81  & 13.28 & 22.95 & 6.40  & 7.20  & 7.29  & 1.00  & 12.39 & 14.20 & 2.74  & 19.12 & 20.81 & 4.39  \\
\rowcolor{blue!5} \textbf{Eyexpert-3.8B} 
& \textbf{43.33} & \textbf{48.67} & \textbf{26.20} & \textbf{26.48} & \textbf{37.09} & \textbf{13.16} & \textbf{16.79} & \textbf{40.71} & \textbf{22.23} & \textbf{19.21} & \textbf{18.08} & \textbf{7.00} & \textbf{40.98} & \textbf{53.54} & \textbf{25.21} \\
\midrule[1.5pt]
\end{tabular}
}
\label{tab:open_additional_results}
\end{table*}

\section{Supplemental Experimental Results}
In this section, we include additional experiments to demonstrate the superiority of Eyecare Kit.
\subsection{Additional Evaluation Metrics}

This section provides a detailed overview of the supplementary evaluation metrics employed in the main experiment. Table~\ref{tab:open_additional_results} outlines the specific metrics used to assess the performance of the open question answering task. Similarly, Table~\ref{tab:report_additional_results} presents the additional evaluation metrics utilized for the report generation task. By including these supplementary evaluations, we aim to provide a more holistic and nuanced understanding of the models' capabilities and limitations in generating free-form textual responses. The results demonstrate that Eyexpert achieved the best performance across both tasks on the supplementary semantic similarity metrics (F1-Rad and BertScore-F1) and the text similarity metrics (BLEU-1 and BLEU-4).

\begin{table*}[htbp]
\centering
\caption{Other evaluation metrics for the report generation task in the main experiment.}
\resizebox{\textwidth}{!}{
\begin{tabular}{lcccccccccccc}
\midrule[1.5pt]
\multirow{2}{*}{\textbf{Model}}& \multicolumn{3}{c}{\textbf{FA-ICGA}} & \multicolumn{3}{c}{\textbf{CT}} & \multicolumn{3}{c}{\textbf{UBM}} \\
\cmidrule(r){2-4} \cmidrule(l){5-7} \cmidrule(l){8-10}
& \textbf{BERTScoreF1} & \textbf{BLEU-1} & \textbf{BLEU-4} & \textbf{BERTScoreF1} & \textbf{BLEU-1} & \textbf{BLEU-4} & \textbf{BERTScoreF1} & \textbf{BLEU-1} & \textbf{BLEU-4}\\
\hline
\rowcolor{gray!5} \multicolumn{10}{c}{\textit{Generalist Models}} \\
\hline
LLaVA-1.5           & 81.12 & 6.06  & 0.23  & 82.27 & 18.44 & 0.84  & 81.01 & 7.57  & 0.12  \\
Qwen2.5-VL           & 84.54 & 26.81 & 0.76  & 84.32 & 16.28 & 2.95  & 81.30 & 8.01  & 0.59  \\
InternVL-2.5        & 82.21 & 7.84  & 0.50  & 83.15 & 17.56 & 0.70  & 81.98 & 9.46  & 0.36  \\
mPLUG-Owl3          & 81.12 & 4.10  & 0.12  & 81.89 & 18.42 & 0.38  & 81.52 & 9.06  & 0.34  \\
Yi-VL              & 80.83 & 7.70  & 0.34  & 83.03 & 19.33 & 1.61  & 80.95 & 7.45  & 0.17  \\
MiniCPM-V2.6      & 81.77 & 9.59  & 0.74  & 82.61 & 17.02 & 1.09  & 81.36 & 8.22  & 0.29  \\
\hline
\rowcolor{gray!5} \multicolumn{10}{c}{\textit{Medical Models}} \\
\hline
LLaVA-Med          & 81.68 & 8.86  & 0.10  & 81.57  & 0.05  & 0.00   & 81.35 & 0.09  & 0.00  \\
MedVLM-R1         & 80.76 & 3.63  & 0.75  & 83.12  & 9.40  & 6  & 81.04  & 7.31  & 1  \\
HealthGPT-M3       & 83.20 & 10.91 & 0.59  & 85.01  & 27.91 & 1.82  & 82.29  & 11.27  & 0.29  \\
\rowcolor{blue!5} \textbf{Eyexpert-3.8B} 
&\textbf{90.12} & \textbf{29.41} & \textbf{2.31} & \textbf{88.36} & \textbf{29.22} & \textbf{2.79} & \textbf{85.70} & \textbf{12.97} & \textbf{0.76} \\
\midrule[1.5pt]
\end{tabular}
}
\label{tab:report_additional_results} 
\end{table*}

\subsection{Eyecare-100K Fine-tuning Ablation Study Results}
In the main text, we only present the experimental performance of EyecareGPT-3.8B before and after fine-tuning on Eyecare-100K. The specific evaluation results are shown in the table~\ref{tab:dataset_ablation}. The results demonstrate a significant improvement in the performance of the fine-tuned EyecareGPT-3.8B across all metrics for each task. Furthermore, the experimental results of EyecareGPT-7B before and after fine-tuning on Eyecare-100K are included in the Appendix, as shown in Table~\ref{tab:7B_ablation}. This supplementary data allows for a more comprehensive evaluation of Eyecare-100K's significant value for ophthalmological AI research.


\input{table/dataset_ablation}

\input{table/data_ablatin_7B}

Findings from the results of the two tables reveal the following:

\noindent \textbf{(i) Fine-tuning Significantly Improves Performance:} Across nearly all tasks and the majority of datasets, both EyecareGPT-3.8B and EyecareGPT-7B demonstrate a substantial performance increase after fine-tuning on the Eyecare-100K dataset (+ Eyecare-100K). This highlights the effectiveness of the Eyecare-100K dataset in adapting these large language models for ophthalmology-specific tasks.

\noindent \textbf{(ii) Larger Models Generally Perform Better:} Comparing the rows before fine-tuning (those not including "+ Eyecare-100K"), EyecareGPT-7B generally exhibits higher initial performance than EyecareGPT-3.8B (Table 10) across most tasks and datasets. This aligns with the common trend that larger language models tend to possess superior zero-shot or few-shot capabilities.

\noindent \textbf{(iii) Fine-tuning Significantly Enhances Large Model Performance:} Despite the higher starting baseline of the 7B model, fine-tuning on Eyecare-100K results in similarly significant absolute gains for this larger model. In many instances, the performance level of the fine-tuned EyecareGPT-7B model considerably surpasses that of the fine-tuned 3.8B model.

\section{Case Study}
\begin{figure*}[t]
    \centering
    \includegraphics[width=0.55\linewidth]{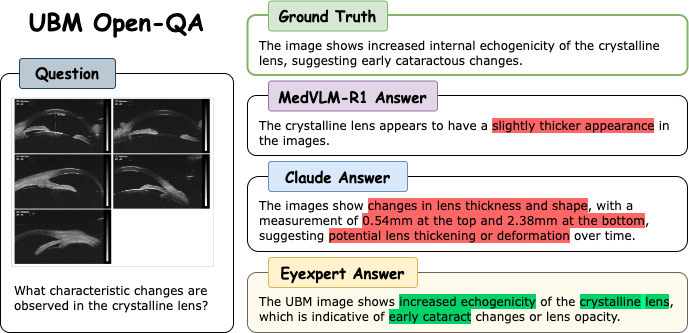}
    \caption{A case of UMB 0pen-QA.}
    \label{fig:ubm-open}
\end{figure*}

\begin{figure*}[t]
    \centering
    \includegraphics[width=0.55\linewidth]{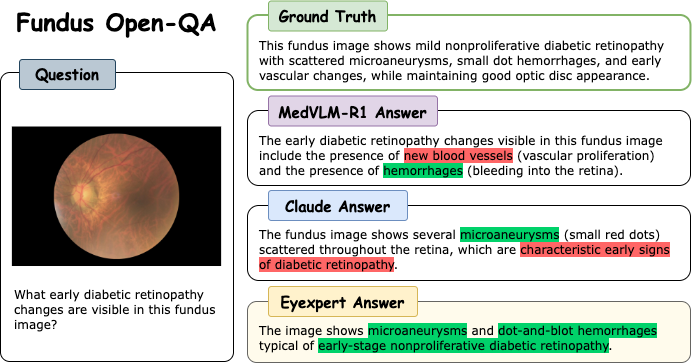}
    \caption{A case of Fundus 0pen-QA.}
    \label{fig:fundus-open}
\end{figure*}

\begin{figure*}[t]
    \centering
    \includegraphics[width=0.55\linewidth]{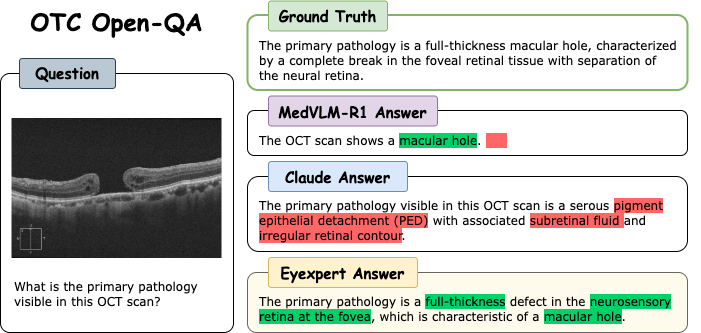}
    \caption{A case of OTC 0pen-QA.}
    \label{fig:otc-open}
\end{figure*}

\begin{figure*}[t]
    \centering
    \includegraphics[width=0.55\linewidth]{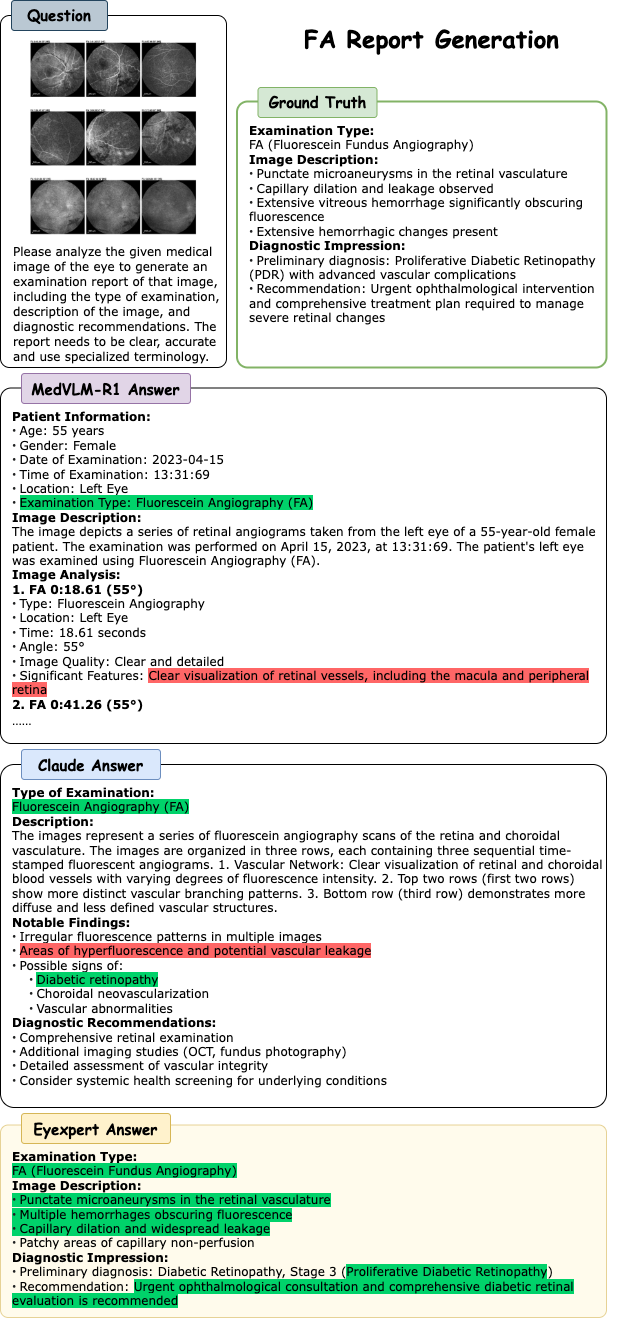}
    \caption{A case of FA Report Generation.}
    \label{fig:fa-report}
\end{figure*}

\begin{figure*}[t]
    \centering
    \includegraphics[width=0.55\linewidth]{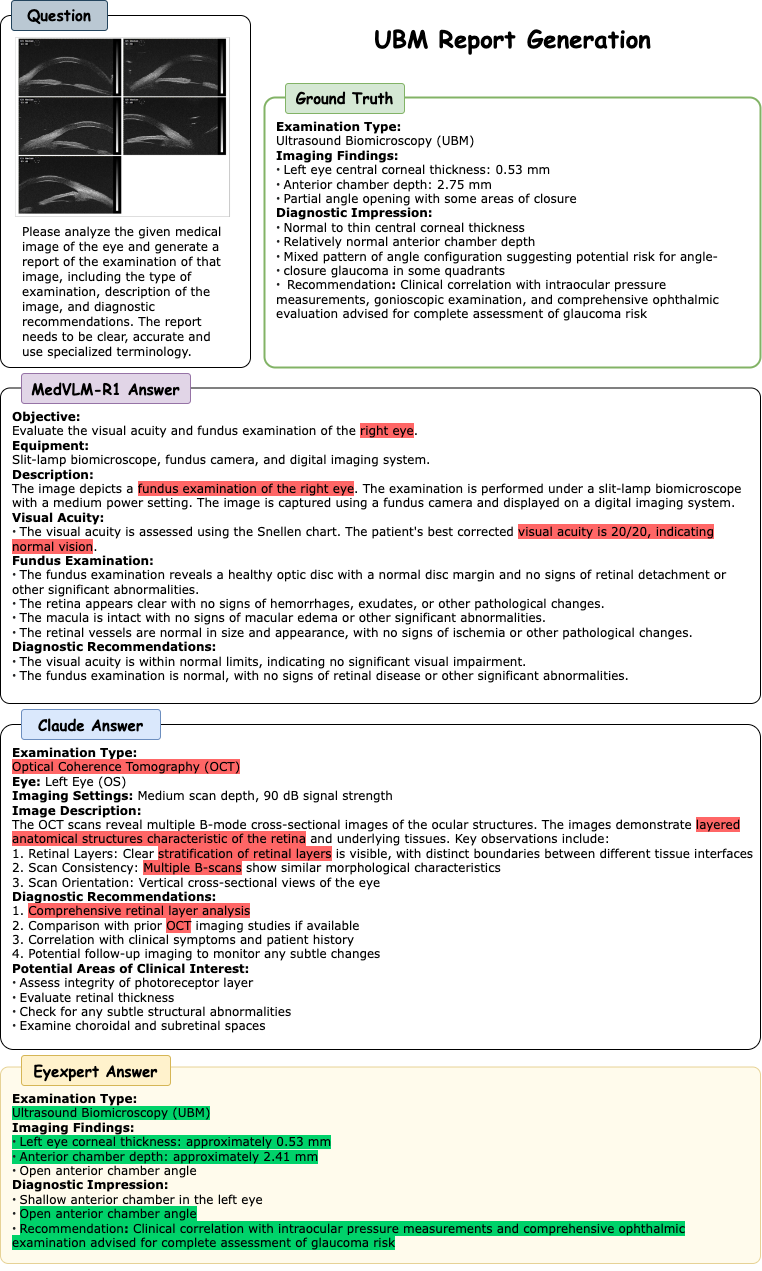}
    \caption{A case of UBM Report Generation.}
    \label{fig:ubm-report}
\end{figure*}

\begin{figure*}[t]
    \centering
    \includegraphics[width=0.55\linewidth]{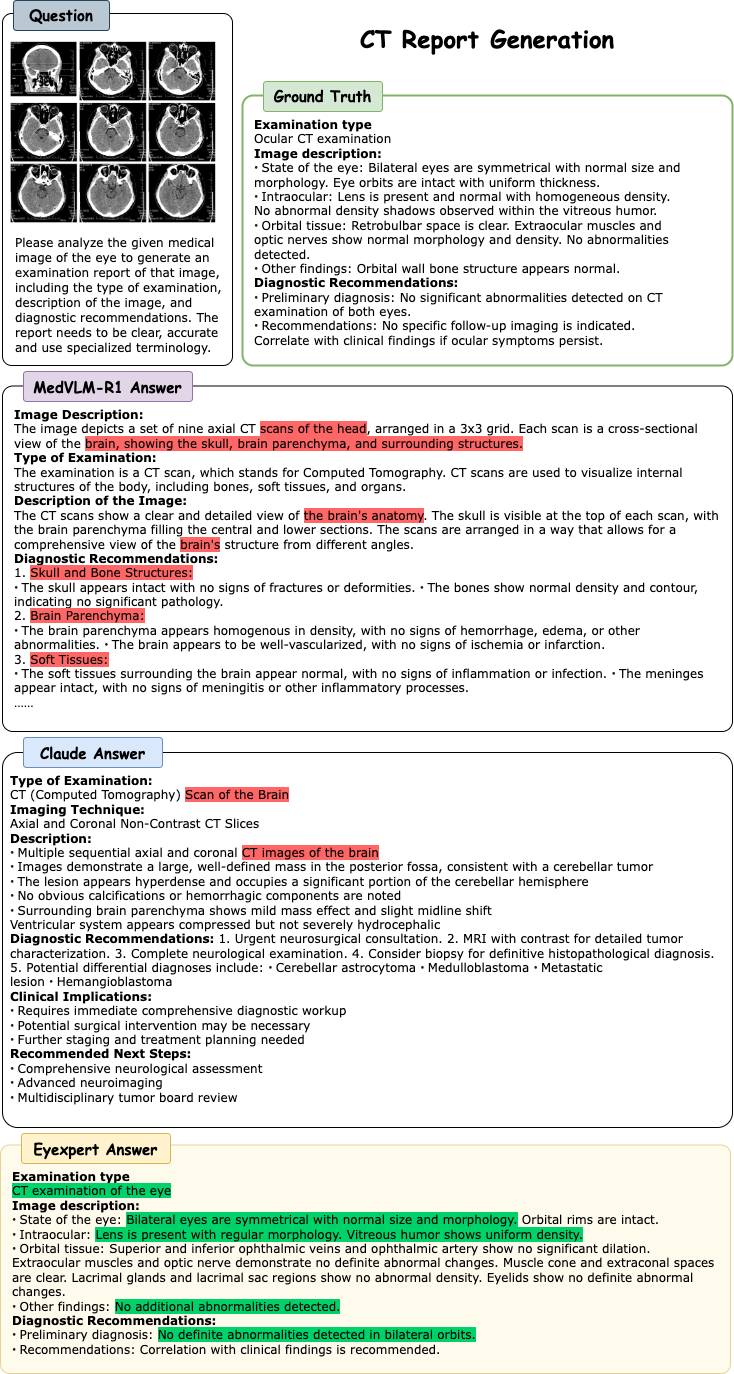}
    \caption{A case of CT Report Generation.}
    \label{fig:ct-report}
\end{figure*}

In this section, we compare the generated answers of our proposed EyecareGPT with those of an open-source medical model (MedVLM-R1) and a closed-source general-purpose model (Claude 3.5). Figures~\ref{fig:ubm-open}, ~\ref{fig:fundus-open}, and ~\ref{fig:otc-open} illustrate the performance of the three models on UBM, CT, and Fundus modalities, respectively, and highlight the differences from the ground truth. Taking Figure~\ref{fig:ubm-open} as an example, our answer is closer to the true answer, demonstrating EyecareGPT's strong understanding of fine-grained diagnostic questions. Figures~\ref{fig:fa-report}, ~\ref{fig:ubm-report}, and ~\ref{fig:ct-report} present report generation examples for Fundus, UBM, and CT modalities. These three figures show that our model can precisely respond to instructions for generating reports, producing well-structured and clear reports that accurately describe abnormal findings in the images.


%% file: table/dataset_ablation.tex
\begin{table*}[t]
\centering
\caption{Comparative Experimental Results of EyecareGPT-3.8B Before and After Fine-tuning on Eyecare-100K.}
\vskip -0.12in

\resizebox{\textwidth}{!}{
\begin{tabular}{cccccccccccc}
\midrule[1.5pt]
\multirow{2}{*}{\textbf{Task}} & \multirow{2}{*}{\textbf{Dataset}} & \multicolumn{7}{c}{\textbf{Eyecare-100K}} & \multicolumn{2}{c}{\textbf{OmniMedVQA}} & \multirow{2}{*}{\textbf{Avg.}}\\
\cmidrule(r){3-9} \cmidrule(l){10-11} 
 & & \textbf{FS.} & \textbf{SL.} & \textbf{OCT} & \textbf{Fundus} & \textbf{FA-ICGA} & \textbf{UBM} & \textbf{CT} & \textbf{OCT} & \textbf{Fundus}\\
\hline
\multirow{2}{*}{Closed-QA}& - & 43.90 & 66.67 & 62.48 & 18.28 & 78.31 & 64.76 & 76.36 & 81.33 & 76.69 & 62.30 \\
 & \textcolor{blue!80}{\textbf{+}} Eyecare-100K & \textbf{60.87} & \textbf{77.03} & \textbf{89.76} & \textbf{75.10} & \textbf{91.43} & \textbf{81.66} & \textbf{85.21} & \textbf{100} & \textbf{100}& \textbf{84.56} \\
\midrule[1pt]
\multirow{2}{*}{\textbf{Task}} & \multirow{2}{*}{\textbf{Dataset}} &\multicolumn{2}{c}{\textbf{OCT}} & \multicolumn{2}{c}{\textbf{Fundus}} & \multicolumn{2}{c}{\textbf{FA-ICGA}} & \multicolumn{2}{c}{\textbf{UBM}} & \multicolumn{2}{c}{\textbf{CT}} \\
\cmidrule(r){3-4} \cmidrule(r){5-6} \cmidrule(r){7-8} \cmidrule(r){9-10} \cmidrule(r){11-12} 
& & \textbf{F1-Bio} & \textbf{Rouge-L} & \textbf{F1-Bio} & \textbf{Rouge-L} & \textbf{F1-Bio} & \textbf{Rouge-L} & \textbf{F1-Bio} & \textbf{Rouge-L} & \textbf{F1-Bio} & \textbf{Rouge-L} \\
\hline
\multirow{2}{*}{Open-QA}& - & 55.53 & 20.10 &  69.80 & 23.43 & 51.79 & 16.47 & 82.12 & 20.08 & 65.22 & 13.60\\
 & \textcolor{blue!80}{\textbf{+}} Eyecare-100K & \textbf{95.53} & \textbf{49.68} & \textbf{90.79} & \textbf{37.03} & \textbf{86.75} & \textbf{49.21} & \textbf{95.78} & \textbf{47.61} & \textbf{83.90} & \textbf{36.11} \\
 \midrule[1pt]
 \end{tabular}}

 \resizebox{\textwidth}{!}{
 \begin{tabular}{cccccccccccccc}
 \multirow{2}{*}{\textbf{Task}} & \multirow{2}{*}{\textbf{Dataset}} & \multicolumn{4}{c}{\textbf{FA-ICGA}} & \multicolumn{4}{c}{\textbf{UBM}} & \multicolumn{4}{c}{\textbf{CT}} \\
\cmidrule(r){3-6} \cmidrule(r){7-10} \cmidrule(r){11-14} 
&  & \textbf{Acc\textsuperscript{GPT}} &  \textbf{Score\textsuperscript{Avg}}& \textbf{F1-Rad} & \textbf{Rouge-L} & \textbf{Acc\textsuperscript{GPT}}  &  \textbf{Score\textsuperscript{Avg}} & \textbf{F1-Rad} & \textbf{Rouge-L} & \textbf{Acc\textsuperscript{GPT}} &  \textbf{Score\textsuperscript{Avg}} & \textbf{F1-Rad} & \textbf{Rouge-L} \\
\hline
\multirow{2}{*}{Report-Gen}& - & 19.21 & 75.35 & 12.78 & 15.19 & 4.51 & 63.41 & 9.36 & 12.19 & 10.71 & 63.93 & 14.94 & 18.92\\
 & \textcolor{blue!80}{\textbf{+}} Eyecare-100K & \textbf{52.62} & \textbf{85.49} & \textbf{25.04} & \textbf{47.91} & \textbf{58.05} & \textbf{79.90} & \textbf{42.83} & \textbf{57.04} & \textbf{50.00} & \textbf{78.64} & \textbf{35.39} & \textbf{42.73} \\
 \midrule[1.5pt]
\end{tabular}
}

\label{tab:dataset_ablation}
\end{table*}

%% file: table/data_ablatin_7B.tex
\begin{table*}[t]
\centering
\caption{Comparative Experimental Results of EyecareGPT-7B Before and After Fine-tuning on Eyecare-100K.}
\vskip -0.12in

\resizebox{\textwidth}{!}{
\begin{tabular}{cccccccccccc}
\midrule[1.5pt]
\multirow{2}{*}{\textbf{Task}} & \multirow{2}{*}{\textbf{Dataset}} & \multicolumn{7}{c}{\textbf{Eyecare-100K}} & \multicolumn{2}{c}{\textbf{OmniMedVQA}} & \multirow{2}{*}{\textbf{Avg.}}\\
\cmidrule(r){3-9} \cmidrule(l){10-11} 
 & & \textbf{FS.} & \textbf{SL.} & \textbf{OCT} & \textbf{Fundus} & \textbf{FA-ICGA} & \textbf{UBM} & \textbf{CT} & \textbf{OCT} & \textbf{Fundus}\\
\hline
\multirow{2}{*}{Closed-QA}& - & 52.17 & 70.33 & 68.82 & 77.36 & 74.71 & 44.78 & 54.93 & 81.93 & 77.36 & 66.93 \\
 & \textcolor{blue!80}{\textbf{+}} Eyecare-100K & \textbf{61.43} & \textbf{77.64} & \textbf{90.09} & \textbf{82.25} & \textbf{92.96} & \textbf{86.78} & \textbf{84.33} & \textbf{99.26} & \textbf{99.56}& \textbf{86.03} \\
\midrule[1pt]
\multirow{2}{*}{\textbf{Task}} & \multirow{2}{*}{\textbf{Dataset}} &\multicolumn{2}{c}{\textbf{OCT}} & \multicolumn{2}{c}{\textbf{Fundus}} & \multicolumn{2}{c}{\textbf{FA-ICGA}} & \multicolumn{2}{c}{\textbf{UBM}} & \multicolumn{2}{c}{\textbf{CT}} \\
\cmidrule(r){3-4} \cmidrule(r){5-6} \cmidrule(r){7-8} \cmidrule(r){9-10} \cmidrule(r){11-12} 
& & \textbf{F1-Bio} & \textbf{Rouge-L} & \textbf{F1-Bio} & \textbf{Rouge-L} & \textbf{F1-Bio} & \textbf{Rouge-L} & \textbf{F1-Bio} & \textbf{Rouge-L} & \textbf{F1-Bio} & \textbf{Rouge-L} \\
\hline
\multirow{2}{*}{Open-QA}& - & 75.84 & 25.91 &  80.24 & 25.12 & 55.01 & 17.01 & 83.14 & 23.66 & 73.17 & 20.28\\
 & \textcolor{blue!80}{\textbf{+}} Eyecare-100K & \textbf{96.26} & \textbf{50.10} & \textbf{90.88} & \textbf{38.13} & \textbf{87.86} & \textbf{51.24} & \textbf{96.60} & \textbf{47.26} & \textbf{87.27} & \textbf{36.70} \\
 \midrule[1pt]
 \end{tabular}}

 \resizebox{\textwidth}{!}{
 \begin{tabular}{cccccccccccccc}
 \multirow{2}{*}{\textbf{Task}} & \multirow{2}{*}{\textbf{Dataset}} & \multicolumn{4}{c}{\textbf{FA-ICGA}} & \multicolumn{4}{c}{\textbf{UBM}} & \multicolumn{4}{c}{\textbf{CT}} \\
\cmidrule(r){3-6} \cmidrule(r){7-10} \cmidrule(r){11-14} 
&  & \textbf{Acc\textsuperscript{GPT}} &  \textbf{Score\textsuperscript{Avg}}& \textbf{F1-Rad} & \textbf{Rouge-L} & \textbf{Acc\textsuperscript{GPT}}  &  \textbf{Score\textsuperscript{Avg}} & \textbf{F1-Rad} & \textbf{Rouge-L} & \textbf{Acc\textsuperscript{GPT}} &  \textbf{Score\textsuperscript{Avg}} & \textbf{F1-Rad} & \textbf{Rouge-L} \\
\hline
\multirow{2}{*}{Report-Gen}& - & 25.33 & 76.02 & 11.36 & 12.48 & 7.27 & 62.83 & 12.79 & 14.88 & 35.71 & 76.00 & 15.16 & 17.38\\
 & \textcolor{blue!80}{\textbf{+}} Eyecare-100K & \textbf{53.91} & \textbf{85.97} & \textbf{26.04} & \textbf{48.32} & \textbf{60.06} & \textbf{80.56} & \textbf{42.98} & \textbf{58.43} & \textbf{52.43} & \textbf{80.71} & \textbf{36.19} & \textbf{43.54} \\
 \midrule[1.5pt]
\end{tabular}
}

\label{tab:7B_ablation}
\end{table*}